%% file: main.tex
\documentclass[10pt,twocolumn,letterpaper]{article}

\usepackage[pagenumbers]{cvpr} % To force page numbers, \eg for an arXiv version

\usepackage[accsupp]{axessibility}
\usepackage{graphicx}
\usepackage{amsmath}
\usepackage{amssymb}
\usepackage{booktabs}
\usepackage{multirow}
\usepackage{diagbox}
\usepackage{bm}
\usepackage{bbm}
\usepackage{mathrsfs}
\usepackage{amsmath}
\usepackage[linesnumbered,ruled]{algorithm2e}

\input{preamble}

\definecolor{cvprblue}{rgb}{0.21,0.49,0.74}
\definecolor{lightgray}{gray}{0.92}
\usepackage[pagebackref,breaklinks,colorlinks,citecolor=cvprblue]{hyperref}

\def\paperID{13584} % *** Enter the Paper ID here
\def\confName{CVPR}
\def\confYear{2024}

\title{Language-Driven Anchors for Zero-Shot Adversarial Robustness}

\author{Xiao Li$^{1}$  \ \ Wei Zhang$^{1}$ \ \ Yining Liu$^{2}$ \ \ Zhanhao Hu$^{1}$ \ \ Bo Zhang$^{1}$ \ \ Xiaolin Hu$^{1, 3}$\thanks{Corresponding author.}\\
$^{1}$Department of Computer Science and Technology, BNRist, Institute for Artificial Intelligence, \\ 
IDG/McGovern Institute for Brain Research, Tsinghua Laboratory of Brain and Intelligence, \\ Tsinghua University, Beijing, China \\
$^{2}$Harbin Institute of Technology, Weihai, China \\
$^{3}$Chinese Institute for Brain Research (CIBR), Beijing, China \\
\tt\small \{lixiao20, zhang-w19, huzhanha17\}@mails.tsinghua.edu.cn \\  \tt\small 22s030184@stu.hit.edu.cn \\  \tt\small \ 	\{dcszb, xlhu\}@mail.tsinghua.edu.cn
}

\begin{document}
\maketitle

\begin{abstract}

Deep Neural Networks (DNNs) are known to be susceptible to adversarial attacks. Previous researches mainly focus on improving adversarial robustness in the fully supervised setting, leaving the challenging domain of zero-shot adversarial robustness an open question. In this work, we investigate this domain by leveraging the recent advances in large vision-language models, such as CLIP, to introduce zero-shot adversarial robustness to DNNs. We propose LAAT, a Language-driven, Anchor-based Adversarial Training strategy. LAAT utilizes the features of a text encoder for each category as fixed \textit{anchors} (normalized feature embeddings) for each category, which are then employed for adversarial training. By leveraging the semantic consistency of the text encoders, LAAT aims to enhance the adversarial robustness of the image model on novel categories. However, naively using text encoders leads to poor results. Through analysis, we identified the issue to be the high cosine similarity between text encoders. We then design an expansion algorithm and an alignment cross-entropy loss to alleviate the problem. Our experimental results demonstrated that LAAT significantly improves zero-shot adversarial robustness over state-of-the-art methods. LAAT has the potential to enhance adversarial robustness by large-scale multimodal models, especially when labeled data is unavailable during training. Code is available at \url{https://github.com/LixiaoTHU/LAAT}.

\end{abstract}

\section{Introduction}
\label{sec:intro}

Adversarial attacks \cite{adv_attack1}, by adding deliberately designed perturbations to inputs, have posed a significant threat to the application of Deep Neural Networks (DNNs) \cite{phy_adv1, phy_adv2}. In response to the threats, various techniques have been proposed to enhance adversarial robustness of DNNs. Among these, Adversarial Training (AT) and its variants \cite{pgdat, mmc, hyper, trades} have shown to be highly effective \cite{adv_attack1, phy_adv1, phy_adv2}. However, applying AT from scratch to novel datasets or categories can be time-consuming. In real-world scenarios, it is common for models to encounter examples from novel categories, known as zero-shot learning (ZSL) setting \cite{zero_shot_e}. If adversarially trained models possess zero-shot capability, it can significantly enhance the scalability and practical utility of adversarially robust models.

\begin{figure}[!t]
  \centering
   \includegraphics[width=\linewidth]{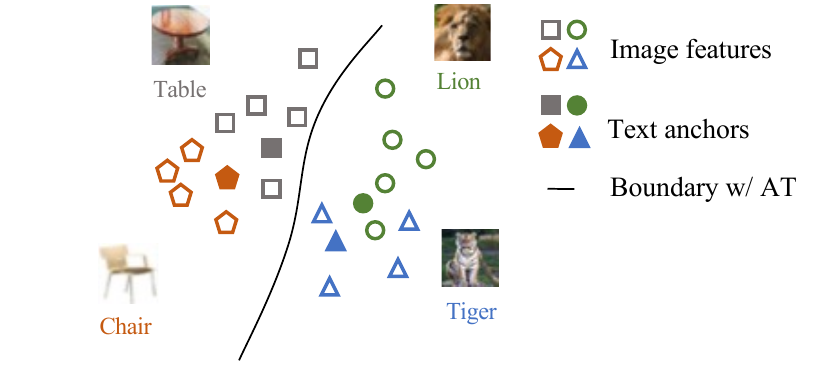}
%	\vspace{-2mm}
   \caption{The illustration of zero-shot ability with LAAT. Different colors of the marks indicate different categories. When a model is adversarially trained with the text anchors of \textit{table} and \textit{lion} (seen categories), it can recognize adversarial examples of the two categories (grey and green). Then due to the text anchors of \textit{chair} and \textit{tiger} (novel categories)  
 being close to those of \textit{table} and \textit{lion}, respectively, the model can also recognize the two novel categories.}
   \label{fig:align}
\end{figure}

Adversarial robustness in the ZSL setting is challenging. Earlier attempts \cite{zerorobust, zeroshot_at} try to combine attribute-based ZSL methods with AT. However, these methods yielded modest improvements and lacked scalability due to the practical difficulties in obtaining attributes. Recently, large-scale vision-language models such as CLIP \cite{clip} have made significant strides and are considered foundational models for various downstream applications. These models, trained on extensive text-image data, demonstrate strong zero-shot capability by using textual descriptions to recognize new visual categories, eliminating the need for explicit reliance on image attributes. In this work, different from previous attribute-based ZSL methods, we explore the utilization of foundation models to introduce zero-shot adversarial robustness to DNNs.

We propose a Language-driven, Anchor-based Adversarial Training strategy, denoted as LAAT, to enhance the zero-shot adversarial robustness. 
Specifically, we use the popular text encoder from vision-language models, specifically CLIP \cite{clip}, to obtain $l_2$ normalized feature embedding (named \textit{anchor} in this paper) for the text (label name) corresponding to image categories. The text encoder exhibits a property called \textit{semantic consistency}, where semantically similar categories are mapped to neighboring anchors in the feature space. Then we utilize the text anchors to supervise the image classification model with AT. After this process, the image model not only obtains adversarial robustness on seen categories but also aligns the image features with the text anchors. During zero-shot inference, by leveraging the semantic consistency between seen and novel categories, the image model can recognize novel categories while maintaining adversarial robustness. \cref{fig:align} illustrates the idea.  We note that one recent work, TeCoA \cite{clip_finetune}, also utilizes the ability of large-scale vision-language models to enhance the zero-shot adversarial robustness. However, TeCoA was only shown effective for extremely tiny adversarial noise (\ie, $\epsilon=1/255$).

As CLIP models are not designed for adversarial robustness, directly applying the obtained anchors from CLIP text encoders to anchor-based AT \cite{mmc, hyper} (refer to \cref{sec:at}) results in poor performance. We attribute it to the high Cosine Similarity (CoS) problem, \ie, the average cosine similarity between CLIP anchors are too high to be directly used in AT. See \cref{sec:problem} for details. This problem has never been discovered in previous work utilizing CLIP models \cite{vild, lseg, clip_finetune, actionclip}. To address it, we propose several techniques. We first introduce an \textit{expansion} algorithm that maps the original CLIP anchors to new anchors with increased distances while maintaining semantic consistency. We then incorporate an Alignment Cross-Entropy (A-CE) loss as the optimization objective. In addition, we use a smoothness loss to enhance the robustness on unseen categories.

The experimental results showed that the models trained with LAAT achieved remarkable zero-shot performance against strong adversarial perturbations, even surpassing the previous state-of-the-art (SOTA) adversarially robust few-shot methods. We also trained several models with LAAT on a large dataset ImageNet-1K \cite{imagenet}. These models showed substantial adversarial robustness across various downstream datasets in the zero-shot setting, significantly outperforming the recent TeCoA method \cite{clip_finetune}. The main contributions can be summarized as follows:
\begin{itemize}
\item We identify the high CoS problem in CLIP anchors for adversarial robustness and propose effective techniques to alleviate it.
\item The models trained by LAAT show substantial adversarial robustness across downstream datasets, suggesting that AT in the ZSL setting could be a promising way to improve the practical usefulness of AT.
\end{itemize}

\section{Related Work}
\label{sec:related}

\subsection{Anchor-based AT}
\label{sec:at}

AT has become one of the most effective strategies to improve adversarial robustness \cite{obs}. Several studies try to improve AT from various aspects \cite{mmc, hyper, trades, rock, semantically, li2023importance}. One series of works focuses on refining cross-entropy loss, which is the most commonly used loss in classification, to obtain better robustness with AT \cite{mmc, hyper}. \citet{mmc} boosts AT by introducing the Max-Mahalanobis center (MMC) loss, a Euclidean loss where there is a fixed feature vector $\mathbf{\mu}_y$ for each category $y$, and the optimization object is $||\mathbf{z} - \mathbf{\mu}_y||_2^2$, where $\mathbf{z}$ is the output feature. \citet{hyper} uses Hypersphere Embedding to boost AT. It normalizes both weights $\mathbf{W}$ and features $\mathbf{z}$ of the output layer, and proposes two losses in the form of $\cos \theta$ and $\theta$, where $\theta$ is the angle between $\frac{\mathbf{W}}{||\mathbf{W}||_2}$ and $\frac{\mathbf{z}}{||\mathbf{z}||_2}$. We call these AT methods \textit{anchor-based} methods as they all used $l_2$ normalized feature embeddings (anchors) for training. Besides, several metric learning works \citep{zhou2022enhancing, mao2019metric} also used a similar paradigm to anchor-based AT for enhancing robustness.

%\vspace{-2mm}
\subsection{Language-driven Recognition.}
Earlier ZSL models exploit auxiliary information, commonly in the form of image attributes, to transfer knowledge between seen and novel categories \cite{zla_3, zla_2}. Recently, language-driven recognition has caught attention in the ZSL field, especially with the recent advance of large-scale pre-trained vision-language models \cite{noises, clip}. CLIP \cite{clip} demonstrated that classic recognition tasks can strongly benefit from millions of raw texts pertaining to images. CLIP uses an independent image encoder and text encoder to perform contrastive learning on 400M image-text pairs. After pre-training, natural language is used to reference visual concepts by computing the CoS between text features and the visual feature, enabling the zero-shot transfer of the model to downstream classification datasets. The basic paradigm of CLIP can be extended to several tasks such as semantic segmentation and object detection \cite{vild, lseg, actionclip}. However, recent works show that CLIP models lack adversarial robustness and can be easily attacked \cite{clip_robust2, clip_finetune, clip_robust3}.

\begin{figure*}[!t]
  \centering
   \includegraphics[width=\linewidth]{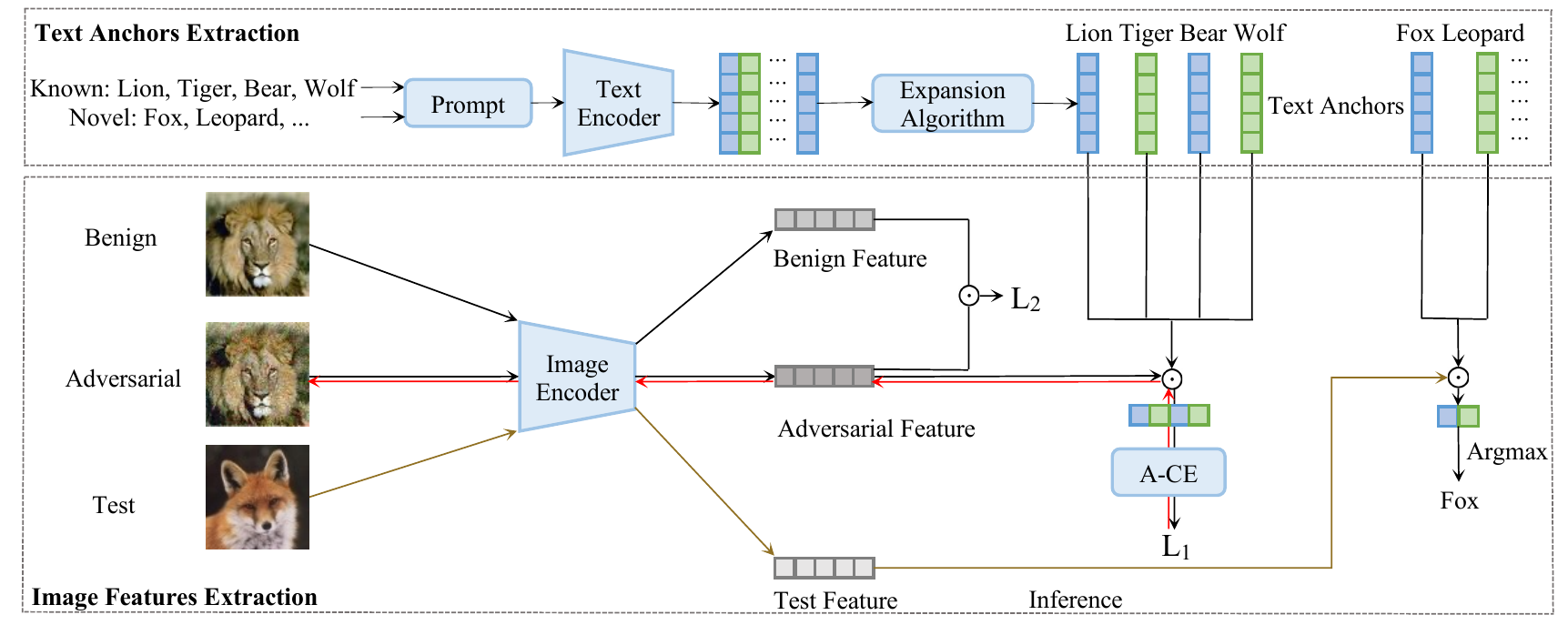}

   \caption{The pipeline of LAAT. Only the image encoder is trainable in the figure. $\odot$ indicates computing the CoS. Red arrows indicate the adversarial example generation process and brown arrows indicate the inference.}
   \label{fig:LAAT}
\end{figure*}

%\vspace{-2mm}
\subsection{Adversarially Robust Zero/few-shot Learning}
An early preprint work \cite{zeroshot_at} combines AT with attribute-based ZSL methods but only gets a mild improvement over non-AT models. In addition, the method is hard to generalize without the attribute annotations. \citet{zerorobust} evaluates the adversarial robustness of classic attribute-based zero-shot models and concludes that adversarial robustness for ZSL field is suffering an immature state. \citet{clip_finetune} proposes several techniques such as visual prompting and fine-tuning to adapt large-scale models, \eg, the image models of CLIP, for zero-shot adversarial robustness, while their method called TeCoA was only shown to be effective under extremely tiny adversarial noise. Different from TeCoA, LAAT only uses the anchors from the CLIP text encoders. 

A closely related area is adversarially robust few-shot learning, which has caught attention recently, too. We also compare LAAT with recent few-shot methods directly to comprehensively evaluate it. AQ \cite{AQ}, R-MAML \cite{AdvMAML}, and ITS-MAML \cite{rethinkingmaml} study meta-learning jointly with AT. GR \cite{basefew} tries to learn a generalizable robust representation by the combination of several AT techniques.

\section{LAAT}
\label{sec:method}

\subsection{Overall Pipeline} 

LAAT first obtains fixed text anchors from the text encoder, then uses them to adversarially supervise the training of image classification model. After AT, the image model will obtain adversarial robustness on seen categories and also align the image features with the text anchors to obtain zero-shot ability (see \cref{fig:align}). The overall pipeline of LAAT is illustrated in \cref{fig:LAAT}. Given the original text anchors, LAAT uses an expansion algorithm to obtain fixed Ground-Truth (GT) anchors suitable for AT, then uses an image encoder to extract both benign and adversarial features. The CoS between adversarial features and GT anchors are maximized by minimizing the A-CE loss ($L_1$). Besides, a smoothness loss ($L_2$) is applied to the adversarial and benign features. We then describe each module of LAAT.

%\vspace{-1mm}
\subsection{Text Anchor Extraction}
\label{sec:clipanchor}
Given the names of $N$ categories in the training set, they are first filled into $N$ sentences with a fixed prompt text, such as ``A photo of \{\}''. After that, these sentences are encoded into $N$ anchors $\{\mathbf{a}_i\}_{i=1}^{N}$ by a fixed pre-trained text encoder, where $\mathbf{a}_i \in \mathbb{R}^n$ and $||\mathbf{a}_i||_{2} = 1$. Many architectures are feasible if they have semantic consistency as mentioned before and in this work, we use the large-scale pre-trained CLIP text encoder. However, CLIP anchors cannot be used directly due to the high CoS problem, as analyzed below. 

%To alleviate it, we first propose an expansion algorithm to enlarge the distances between anchors (reducing the CoS).

%\vspace{-1mm}
\subsection{High Cosine Similarity Problem}
\label{sec:problem}
\begin{figure}[t]
  \centering
   \begin{subfigure}[b]{0.49\linewidth}
         \centering
         \includegraphics[width=\linewidth]{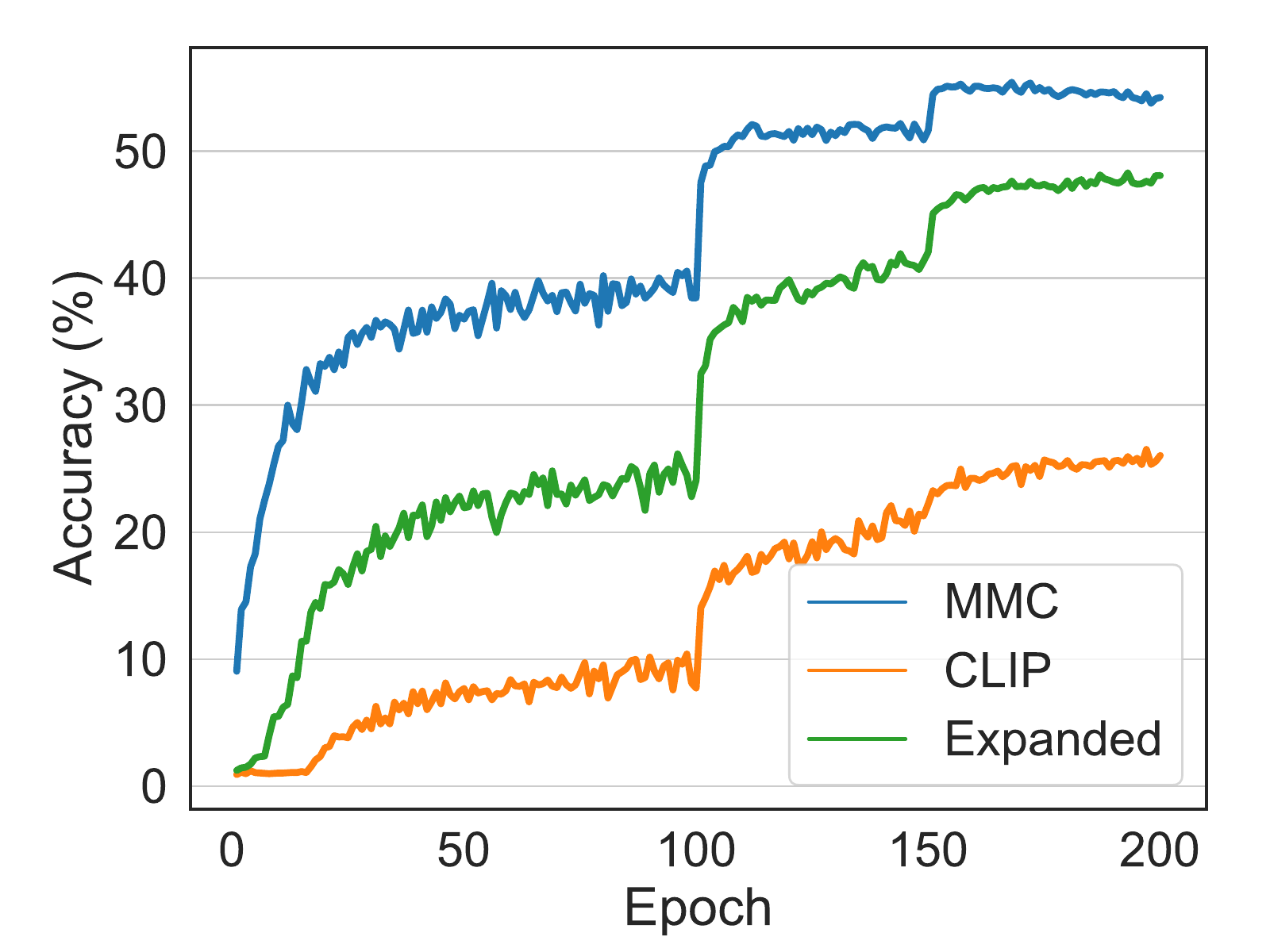}
         \caption{Clean accuracy.}
     \end{subfigure}
     \hfill
     \begin{subfigure}[b]{0.49\linewidth}
         \centering
         \includegraphics[width=\linewidth]{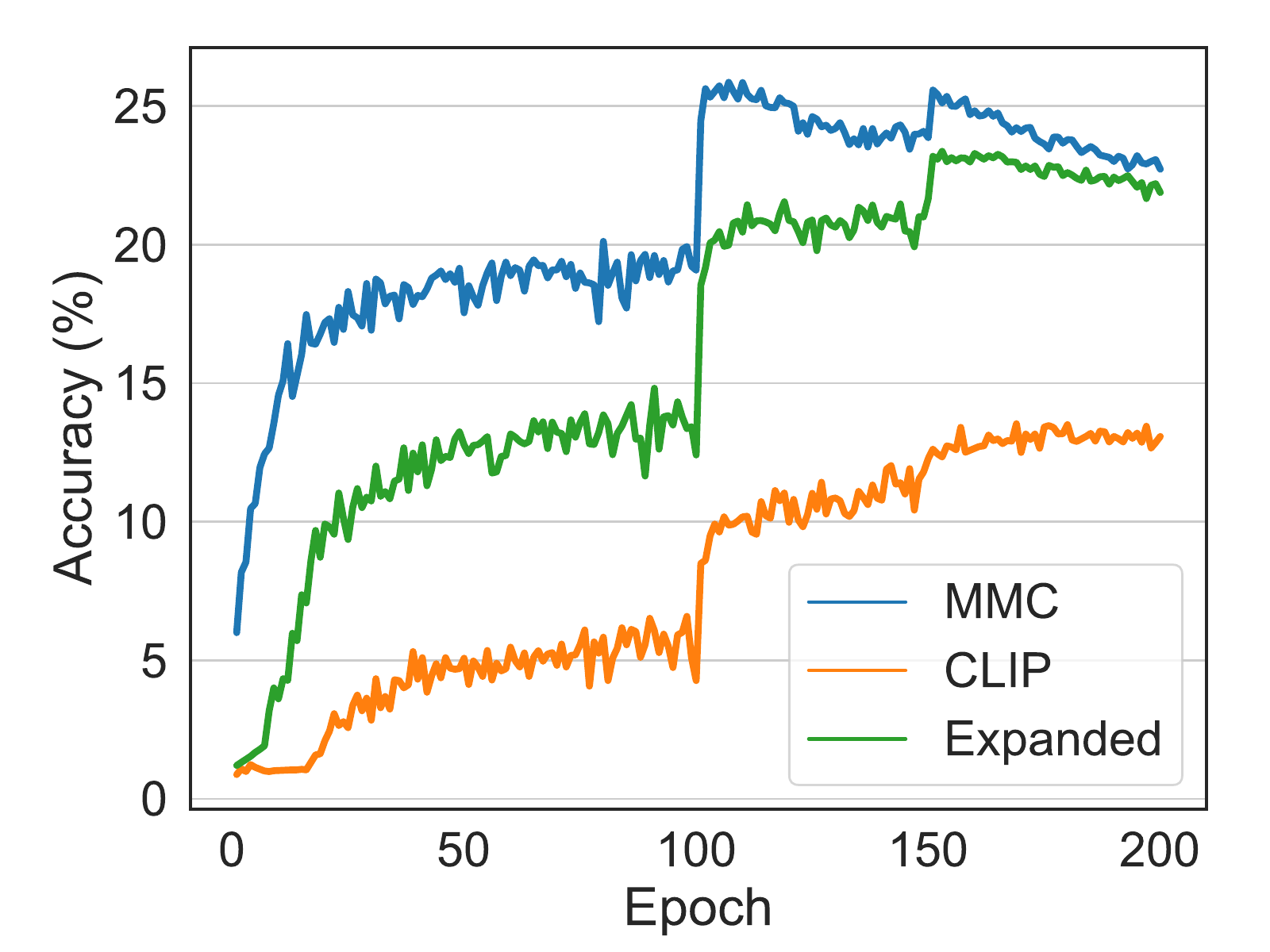}
         \caption{Robust accuracy.}
     \end{subfigure}
%   \vspace{-2mm}
   \caption{Learning curves of AT supervised by $\cos \theta$ with fixed anchors generated from MMC method, CLIP text encoder, and the expansion algorithm (see \cref{sec:expansion}).}
   \label{fig:curve}
\end{figure}

We first make an intriguing observation that directly using CLIP anchors for anchor-based AT poses significant challenges in achieving convergence. This phenomenon is universal among different CLIP text encoders and different models. To illustrate this, we adversarially trained a ResNet18 \cite{resnet12} on Cifar100 \cite{cifar10} by maximizing the $\cos \theta$ objective \cite{hyper}, \ie, the CoS between the model output $\mathbf{z}$ and the GT anchor $\mathbf{a}_y$ \cite{hyper}, which is computed as $\frac{\mathbf{z}^T\mathbf{a}_y}{||\mathbf{z}||_2}$ (note that $||\mathbf{a}_y||_2=1$). The experimental details can be found in \cref{sec:experimental_cifar100}. \cref{fig:curve} shows the learning curves with CLIP anchors (the orange curves). For comparison, here we also show the learning curves with MMC \cite{mmc} anchors (the blue curves) in the same training setting. MMC anchors are pre-computed by maximizing the distance between anchors \cite{mmc} and thus the average CoS between MMC anchors is close to 0 (\eg, -0.01 on Cifar100). The accuracy with CLIP anchors increased quite slowly. Even after 200 epoch training, the accuracy was far lower than that of MMC anchors (note that zero-shot recognition cannot be achieved with MMC anchors for the lack of semantic consistency).

The stark discrepancy between the two learning curves motivates us to analyze the essential difference between the two sets of anchors. Inspired by the anchor generation process of MMC, we calculated the CoS between CLIP anchors and found them to be considerably larger than that of MMC anchors (-0.01). The first row of \cref{tab:cosine} presents the average CoS between anchors from Cifar100 categories generated by different CLIP models. Note that the high CoS between CLIP anchors is irrelevant to the prompt text. We found that even using several random sentences from Wikipedia to encode anchors by CLIP text encoders, the average CoS remained significantly large.

%(\eg, $\ge 0.58$ in our experiment). 

Although high CoS seems to have no effect on downstream tasks on clean images \cite{vild, lseg, actionclip}, we discovered that AT was remarkably affected because higher CoS implies that the anchors are closer to the decision boundary, which could leave more room for adversarial examples \cite{mmc, trades}. This problem may also explain why TeCoA \cite{clip_finetune}, which also used the text anchors during adaptation, failed to convert CLIP image models to adversarially robust models against strong perturbations. Importantly, random or MMC anchors cannot mitigate this issue as they lack the semantic consistency provided by CLIP anchors and thus cannot be used in the ZSL setting. Therefore, the micro designs in our LAAT approach primarily aim to address the challenges associated with CLIP anchors.

%We also investigated the linear output layer of a ResNet18 with standard AT, where the learnable weights for each category can be regarded as learnable unnormalized anchors $\{\mathbf{w}_i\}_{i=1}^{N}$. We calculated the CoS between $\{\mathbf{w}_i\}$, and found that the average CoS between $\{\mathbf{w}_i\}$, like that of the MMC anchors, was also quite close to $0$.

\begin{table}[!t]
  \centering
%  \vspace{-2mm}
  \small
%  \resizebox{\linewidth}{36mm}{
%	\renewcommand{\arraystretch}{1.15}{
 \setlength{\tabcolsep}{2.5pt}
  {
    \begin{tabular}{c|ccccc}
    \toprule
     Type & RN50x4 & RN50x16 & ViT-B/32 &  ViT-B/16 &  ViT-L/14 \\
     \midrule
%	\hline
     Original & $0.700$ & $0.710$ & $0.779$ & $0.761$ & $0.746$ \\ 
     Expanded & $0.238$ & $0.253$ & $0.199$ & $0.195$ & $0.222$ \\
     \bottomrule
    \end{tabular}
    
    }
%   }
%	\vspace{-2mm}
   \caption{Average CoS between anchors before and after expansion algorithm (see \cref{sec:expansion}) from Cifar100 categories generated by different CLIP text encoders.}
  \label{tab:cosine}
\end{table}

%\vspace{-1mm}
\subsection{Expansion Algorithm}
\label{sec:expansion}

\begin{figure}[t]
  \centering
   \includegraphics[width=0.98\linewidth]{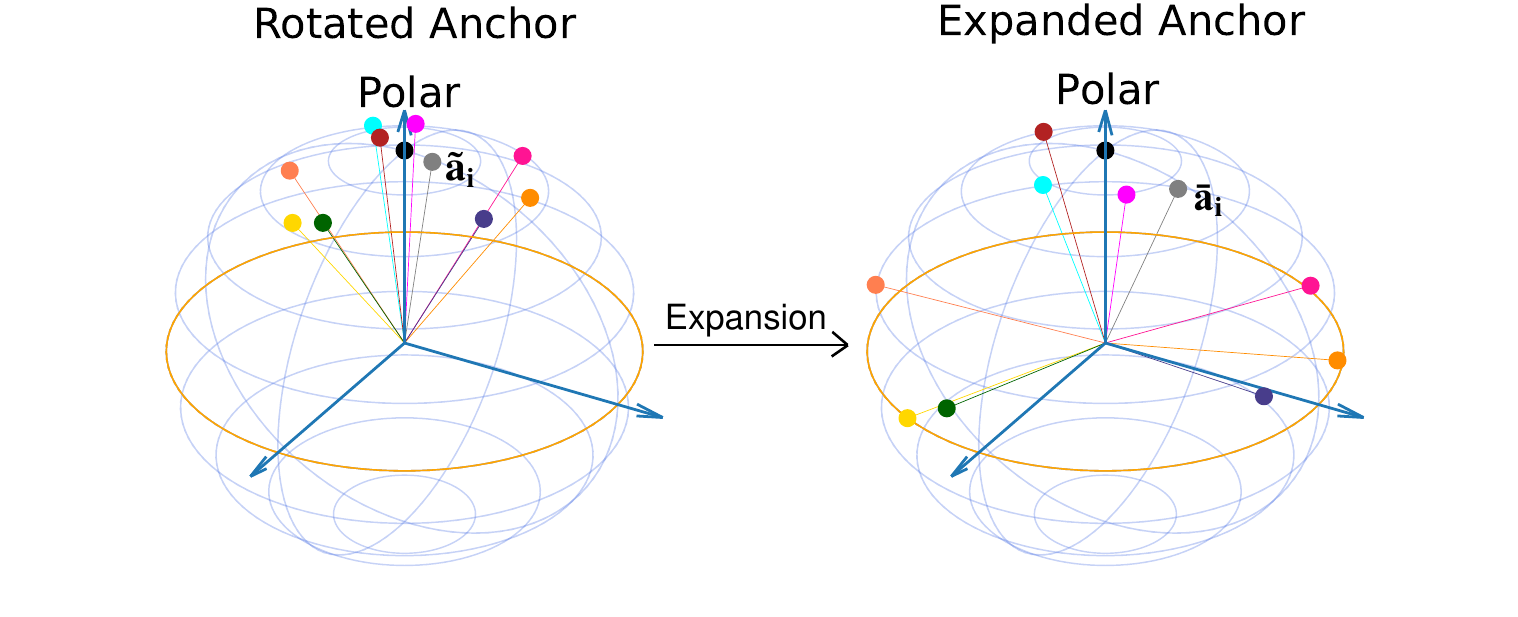}
   \vspace{-1mm}
   \caption{An illustration of the expansion operation in 3D space.}
   \label{fig:sphere}
\end{figure}

We first propose an expansion algorithm to increase the distances between anchors (reducing the CoS). The expansion algorithm can remap the anchors to increase the distances between anchors while preserving the semantic consistency between them. High CoS means the normalized anchors $\{\mathbf{a}_i\}_{i=1}^{N}$ are dispersed over a cluster of the unit hyper-sphere (see \cref{fig:sphere}). As we expect the remapped anchors to be also on the unit hyper-sphere, the expansion algorithm should be designed under spherical coordinates. Let us recap the transformation between $n$-dimensional spherical coordinate system and Cartesian coordinates. Suppose $(x^{(1)}, x^{(2)}, ..., x^{(n)})$ is the Cartesian coordinate, we can compute them by:
\begin{equation}
\label{eq:coord}
\begin{aligned}
&    x^{(1)} = r\cos\phi^{(1)}; \quad x^{(2)} = r\sin\phi^{(1)} \cos\phi^{(2)}; \quad \cdots; \\
&   x^{(n-1)} = r\sin\phi^{(1)}\cdots\sin\phi^{(n-2)}\cos\phi^{(n-1)}; \\
&    x^{(n)} = r\sin\phi^{(1)}\cdots\sin\phi^{(n-2)}\sin\phi^{(n-1)},
\end{aligned}
\end{equation}
where $r$ is the radial coordinate and $\phi^{(1)}, \cdots, \phi^{(n-1)}$ are angular coordinates \cite{coord}. Here $\phi^{(1)}$ is the polar angle (taking the three-dimensional spherical coordinate system like longitude and latitude system as an example, the polar angle is the angle with respect to the $z$ axis). 
If the anchors are clustered around the polar, \ie, $\phi_0:=\max_{i}{\phi_{i}^{(1)}} < \frac{\pi}{2}$, where $\phi_{i}^{(1)}$ denotes the polar angle of anchor $\mathbf{a}_i$, a natural expansion method is to enlarge the polar angle $\phi_{i}^{(1)}$ to $\frac{\pi}{2} \cdot \phi_{i}^{(1)}/\phi_0$ and keep the other angular coordinates unchanged. Then the cluster can be expanded to the whole hemisphere. Note that we cannot expand them larger than $\frac{\pi}{2}$ as anchors may be close to each other from another hemisphere, which may hurt the semantic consistency of anchors.

%  range over  $[0,\pi]$, and $\phi_{n-1}$ is an angle range over $[0,2\pi]$

Thus, given the \textit{original anchor} $\mathbf{a}_i$, the first step of the expansion is to calculate a rotation matrix $\mathbf{R}$ to transform the center $\mathbf{v}$ of the clustered $\mathbf{a}_i$'s ($\mathbf{v}:= \sum_{i=1}^{N}{\mathbf{a}_i} / ||\sum_{i=1}^{N}{\mathbf{a}_i}||_2$) to the polar $\mathbf{p} = [1, 0, \cdots, 0]^T$, and the \textit{rotated anchor} $\tilde{\mathbf{a}}_i$ is computed by: $\tilde{\mathbf{a}}_i = \mathbf{R}\mathbf{a}_i$. Note that computing $\mathbf{R}$ between two unit vectors has several ways and we use the \citet{rotate} algorithm. As $\tilde{\mathbf{a}}_i$'s are clustered around the polar $\mathbf{p}$, we then enlarge the polar angle of $\tilde{\mathbf{a}}_i$ as described above. Denoting the $j$-th element of $\mathbf{a}_i$ as $a_{i}^{(j)}$, the largest polar angle between $\tilde{\mathbf{a}}_i$'s and $\mathbf{p}$ can be computed as $\phi_0 = \max_{1 \leq i \leq N}{\{\arccos{\tilde{a}_{i}^{(1)}}\}}$. According to \cref{eq:coord}, the Cartesian coordinates of \textit{expanded anchor} $\bar{\mathbf{a}}_i$ is given by:
\begin{equation}
\begin{split}
& \bar{a}_{i}^{(1)} = \cos{\bar{\phi} _{i}^{(1)}}; \\
& \bar{a}_{i}^{(j)} = \tilde{a}_{i}^{(j)} \cdot \sin{\bar{\phi}_{i}^{(1)}} / \sin{\tilde{\phi}_{i}^{(1)}}, \quad  j = 2,\cdots,n,
\end{split}
\end{equation}
where  $\tilde{\phi}_{i}^{(1)} = \arccos \tilde{a}_{i}^{(1)}$ denotes the polar angle of each $\tilde{\mathbf{a}}_i$ and $\bar{\phi}_{i}^{(1)} = \frac{\pi}{2} \cdot \tilde{\phi}_{i}^{(1)}/\phi_0$ denotes the expanded polar angle.  \cref{fig:sphere} illustrates this operation and the detailed derivation is shown in \cref{sec:derivation_expand_anchor}. 
%where $\mathbf{a}_{i,j}$ denotes the $j$-th element of $\mathbf{a}_i$. 
In the end, we use the inverse of $\mathbf{R}$ to map the expanded anchor $\bar{\mathbf{a}}_{i}$ to the original locations and obtain the \textit{final anchor}: $\hat{\mathbf{a}}_i = \mathbf{R^T}\bar{\mathbf{a}}_i$. The pseudo-code of the whole expansion algorithm is provided in \cref{sec:pseudo_code}. As shown, the expansion algorithm can increase the distances between different anchors while maintaining the semantic consistency of the original CLIP anchors (close anchors are still close after expansion).

The second row of \cref{tab:cosine} shows the average CoS between anchors after the expansion algorithm. Compared with the original CLIP anchors, the average CoS between the expanded CLIP anchors is greatly reduced. \cref{fig:curve} shows the learning curves with the expanded CLIP anchors (the green curves). The final robust accuracy (\cref{fig:curve}, right) is quite close to that of MMC anchors.

 \begin{table*}[!t]
  \centering
%  \vspace{-2mm}
  \small
%  \resizebox{\linewidth}{36mm}{
%\renewcommand{\arraystretch}{1.15}{
%  \setlength{\tabcolsep}{3.8pt}
  {
    \begin{tabular}{c|c|c|c|cccc}
%    \toprule
    Dataset & Method  & Setting & Clean & FGSM & PGD & CW & AA \\
    
     \hline
%	\midrule
	\multirow{5}*{CIFAR-FS} & AQ \cite{AQ} & 1-shot   & $54.31 \pm 0.50$ & $32.47 \pm 0.47$     & $30.41 \pm 0.46$     & $30.14 \pm 0.46$ & $27.09 \pm 2.07$ \\
   & R-MAML \cite{AdvMAML} & 1-shot   & $38.53 \pm 0.47$ & $26.76 \pm 0.44$     & $26.21 \pm 0.43$     & $29.76 \pm 0.46$ & $16.67 \pm 5.48$ \\
   & GR \cite{basefew} & 1-shot   & $45.27 \pm 0.49$ & $39.60 \pm 0.46$     & $38.03 \pm 0.46$     & $37.00 \pm 0.46$ & - \\
	\cline{2-8}
	& LAAT & zero-shot & $\textbf{55.60}\pm0.46$ & $\textbf{41.17}\pm0.44$ & $\textbf{40.12}\pm0.44$ & $\textbf{37.35}\pm0.44$ & $\textbf{36.45}\pm1.90$ \\
%	\hline
	& LAAT & 1-shot & $ \textbf{57.88}\pm0.45$ & $\textbf{43.73}\pm0.44$ & $\textbf{42.74}\pm0.44$ & $\textbf{40.10}\pm0.45$ & $\textbf{39.23}\pm2.12$ \\
	
%	\midrule
%	\midrule
%	\hline
	\hline
	\multirow{5}*{MiniImageNet} & AQ \cite{AQ} & 1-shot   & $39.17 \pm 0.40$ & $22.53 \pm 0.35$     & $20.95 \pm 0.34$     & $19.33 \pm 0.38$ & $17.85 \pm 1.41$ \\
	& R-MAML \cite{AdvMAML} & 1-shot   & $36.18 \pm 0.36$ & $23.85 \pm 0.46$ & $21.74 \pm 0.34$ & $19.68 \pm 0.40$ &  $13.69 \pm 3.95$ \\
	
   & GR \cite{basefew} & 1-shot   & $36.14 \pm 0.45$ & $29.23 \pm 0.33$     & $27.57 \pm 0.38$     & $26.61 \pm 0.33$ & - \\
%   \cmidrule{2-8}
	\cline{2-8}
   & LAAT & zero-shot & $\textbf{46.97}\pm0.35$ & $\textbf{30.69}\pm0.30$ & $\textbf{29.27}\pm0.31$ & $\textbf{27.91}\pm0.29$ & $\textbf{25.24}\pm1.42$ \\
   & LAAT & 1-shot & $\textbf{47.37}\pm0.35$ & $\textbf{31.55}\pm0.31$ & $\textbf{29.93}\pm0.31$ & $\textbf{28.19}\pm0.29$ & $\textbf{25.75}\pm1.45$ \\
%	\bottomrule
    \end{tabular}
    
    }
%    }
%     }
%  \vspace{-2mm}
  \caption{The comparison of LAAT in 5-way zero-shot setting and previous works in 5-way 1-shot setting under $\epsilon=\mathbf{8}/255$. The 95\% confidence interval was reported. Here all methods used Conv4-512 as the image model. The accuracy of GR under AA is not shown as it was not reported and no code was released. We also list the performance of LAAT in the 1-shot setting with image-text blended anchors (discussed in \cref{sec:ztof}). The results higher than previous methods in each column are in \textbf{bold}.}
  \label{tab:cifarfs}
\end{table*}

 \begin{table*}[!t]
  \centering
  \small
%  \resizebox{\linewidth}{36mm}{
%\renewcommand{\arraystretch}{1.15}{
  \setlength{\tabcolsep}{3pt}
  {
    \begin{tabular}{c|c|c|cccccccccccc}
%    \toprule
    \multirow{2}*{Method} & \multirow{2}*{Model} & \multirow{2}*{Training} & Cifar100  & aPY & AwA2 & COCO Obj. & STL10 & OxfordPet & DTD & Caltech101 & Caltech256 & SUN \\
    & & & 100 & 32 & 50 & 80 & 10 & 37 & 47 & 101 & 257 & 397 \\ 
%    \midrule
	\hline
	\multirow{2}*{TeCoA} & ViT-B/32$^\star$ & $\epsilon = 1/255$ & $0.00$ & $4.39$ & $0.88$ & $1.07$ & $0.98$ & $0.00$ & $1.43$ & $1.66$ & $1.33$ & $0.00$ \\
    \multirow{2}*{\cite{clip_finetune}}  & ViT-B/32 & $\epsilon = 8/255$ & $0.00$ & $5.27$ & $1.45$ & $1.33$ & $1.46$ & $0.28$ & $1.89$ & $2.97$ & $2.36$ & $0.14$ \\
%    \multirow{2}*{\cite{clip_finetune}} & ViT-B/16 & $\epsilon = 1/255$ & $0.10$ & $2.44$ & $0.29$ & $0.29$ & $0.98$ & $0.00$ & $0.78$ & $3.32$ & $1.07$ & $0.00$ \\
    & ViT-B/16 & $\epsilon = 8/255$ & $0.20$ & $6.45$ & $1.95$ & $1.56$ & $1.66$ & $0.49$ & $2.25$ & $7.03$ & $3.81$ & $0.29$ \\
    \hline
    \multirow{2}*{LAAT} &  ViT-B/32 & $\epsilon = 8/255$ & $6.64$ & $24.32$ & $23.34$ & $18.56$ & $41.12$ & $20.31$ & $4.59$ & $34.43$ & $21.39$ & $4.30$ \\
    	
    & ViT-B/16 & $\epsilon = 8/255$ & $\textbf{7.32}$ & $\textbf{34.97}$ & $\textbf{26.21}$ & $\textbf{20.61}$ & $\textbf{49.41}$ & $\textbf{28.81}$ & $\textbf{4.79}$ & $\textbf{37.89}$ & $\textbf{24.02}$ & $\textbf{5.18}$ \\
		
%     & XCiT-S12/16 & $6.93$ & $32.75$ & $24.00$ & $19.46$ & $45.41$ & $21.78$ & $\textbf{5.66}$ & $36.04$ & $21.68$ & $4.79$ \\
%     & ConvNeXt-T & \\
%    \bottomrule
    
    \end{tabular}
    
    }
%    }
%     }
%  \vspace{-2mm}
  \caption{Adversarial accuracy $(\%)$ of models trained on ImageNet-1K on ten downstream datasets in GZSL setting. The number of categories is shown below the dataset name.  $^\star$ indicates the results of the originally released checkpoint of TeCoA with the training perturbation $\epsilon = 1/255$. The robustness was all evaluated by PGD-20  under $\epsilon = \mathbf{8} /255$. }
  \label{tab:pretrain}
\end{table*}

%\vspace{-1mm}
\subsection{Supervision Objective}
\label{sec:objective}

We propose to improve the anchor-based AT methods further by introducing an A-CE loss into the supervision objective. The A-CE loss is a composition of CE loss and the anchor alignment process (the $\cos \theta$ objective), which can also be seen as a special case of InfoNCE \citep{infonce} with using $l_2$ normalized features. This could be confusing considering that previous works have shown that CE is inferior to other anchor-based objectives such as the $\theta$ and $\cos \theta$ \cite{hyper} in the standard AT setting (see \cref{sec:at}). However, we find that with the high CoS problem, A-CE can boost the performance of anchor-based AT methods in that the softmax function of A-CE could make a relaxation on the anchor alignment objective of the original anchor-based AT methods. Assuming that the image encoder $f_\Theta(\cdot)$ outputs normalized features $\mathbf{z} = f_\Theta(\mathbf{x}) \in \mathbb{R}^n$ and  $||\mathbf{z}||_2 = 1$ and taking the $\cos \theta$ objective as an example, the $\cos \theta$ objective is to maximize $\mathbf{z}^T\hat{\mathbf{a}}_y$, where $\hat{\mathbf{a}}_y$ denotes the GT anchor. It encourages the visual feature's CoS to be as close as possible to the GT anchor, while the A-CE loss, by putting $\cos \theta$ into the softmax function, only encourages the visual feature's CoS with GT to be larger than those with other anchors. With A-CE supervision, the output feature could be optimized to be far away from all anchors, but it can still be correctly classified as long as it is closest to the GT anchor. This will be discussed further in \cref{sec:ablation}. Overall, the supervision objective is: 
\begin{equation}
\begin{aligned}
	L_1 = \mathbb{E}_{(\mathbf{x}, y)}\left[-\log {\frac{\exp(f_\Theta(\mathbf{x} + \mathbf{\delta})^T\hat{\mathbf{a}}_y)}{\sum_{i=1}^N \exp(f_\Theta(\mathbf{x} + \mathbf{\delta})^T \hat{\mathbf{a}}_i)}}\right],
\end{aligned}
\end{equation}
where $\mathbf{\delta}$ is adversarial perturbation generated by $\frac{\partial L_1}{\partial \mathbf{x}}$. Unlike previous works \cite{clip} on standard training, we do not use a temperature parameter $\tau$ to scale the CoS. This is discussed in \cref{sec:ablation_tau}.

% indicate that this may be harmful under AT instead.

%  $\mathbf{z}_{\rm adv} = f_\Theta(\mathbf{x} + \mathbf{\delta}) \in \mathbb{R}^n$, and $||\mathbf{z}_{\rm adv}||_2 = 1$.

%\vspace{-1mm}
\subsection{Smoothness Loss}
Since we prefer the adversarial robustness of the model in the zero-shot transfer setting, we additionally introduce a smoothing loss, encouraging the CoS between adversarial features to be similar to the benign features of an image:
\begin{equation}
\begin{aligned}
	L_2 = \mathbb{E}_{(\mathbf{x}, y)}\left[ - f_\Theta(\mathbf{x})^T f_\Theta(\mathbf{x}+\mathbf{\delta}) \right].
\end{aligned}
\end{equation}
The smoothing loss is independent of the category labels of training examples. It can be expected to improve the adversarial robustness on novel categories. Finally, the whole loss in the training stage is given by $L = L_1 + \alpha L_2$, where $\alpha$ is a hyper-parameter. 

\subsection{Performing Zero-Shot Recognition}

When performing zero-shot recognition on novel categories, the image features of a test image are generated by the image encoder, and the anchors for novel categories are generated by the text encoder and the expansion algorithm. Note that the internal parameters of the expansion algorithm, such as $\phi_0$, are pre-computed using anchors from the training categories. Finally, the zero-shot prediction is the category whose text anchor exhibits the highest CoS with the image features.

\section{Experiments}
\label{sec:experiment}

%we first applied LAAT to two popular few-shot benchmark datasets and compared our method with several adversarially robust few-shot methods \cite{AQ, AdvMAML, basefew} directly.

\subsection{Experimental Setup}
\paragraph{Baseline.}
We first consider the standard ZSL setting. Due to the lack of adversarially robust methods in this setting, we compared LAAT with several robust few-shot methods directly: AQ \cite{AQ}, R-MAML \cite{AdvMAML}, and GR \cite{basefew}. ITS-MAML \cite{rethinkingmaml} was not compared as it used a different attack setting and no code was released. Note that zero-shot is harder than few-shot as the latter can have access to a few examples from the novel categories.

We then considered a more pragmatic version of ZSL, which requires recognizing examples from both seen and novel categories. This is called the generalized zero-shot learning (GZSL) setting \cite{gzsl}. We trained several models with LAAT on ImageNet-1K and evaluated them directly on ten downstream datasets. In this setting, we mainly compare LAAT with TeCoA \cite{clip_finetune}. We did not compare with \citet{zerorobust} and \citet{zeroshot_at} as they relied on image attributes and were unsuitable for this setting.

% It contains 600 images of 32 $\times$ 32 size per category.
\vspace{-2mm}
\paragraph{Datasets.} 

When comparing with few-shot methods, we used two popular few-shot benchmarks CIFAR-FS \cite{cifarfs} and MiniImageNet \cite{MiniImageNet}. CIFAR-FS, a variant of Cifar100 \cite{cifar10}, contains 64, 16, and 20 categories for training, validation, and testing, respectively. MiniImageNet, a subset of ImageNet-1k \cite{imagenet}, contains 600 images of 84 $\times$ 84 size per category. We used the same dataset splitting setting as previous works \cite{basefew, AQ, AdvMAML}, for training and testing.

In the GZSL setting, we tested models trained on ImageNet-1K on ten downstream datasets such as AwA2 \cite{awa2} and aPY \cite{aPY}. Note that all of these datasets include several novel categories that have never appeared in ImageNet-1K, \eg, \textit{statue} and \textit{ass} in aPY \cite{aPY}, \textit{blue whale} and \textit{dolphin} in AwA2 \cite{awa2}. More introduction to the datasets is shown in the \cref{sec:downsteam_datasets}.

\vspace{-2mm}
\paragraph{Implementation details.} Unless otherwise stated, we used the CLIP ViT-B/16 text encoder and the prompt text was set as ``This is a photo of a \{\}''. $\alpha$ was set to 3.

In the standard ZSL setting, we followed previous few-shot works \cite{basefew, AdvMAML} and used the widely applied model Conv4-512 \cite{conv4_512} as the image model. The training recipe strictly followed GR \cite{basefew}. We performed AT by generating adversarial examples via PGD method with maximum adversarial perturbation $\epsilon=8/255$ in $l_{\infty}$-norm bound, with $T = 7$ iterative steps (step size $s = 2/255$).

In the GZSL setting, we trained two image models ViT-B/32 and ViT-B/16 \cite{transformer} with LAAT on the large ImageNet-1K. The training basically followed the recipe proposed by \citet{recipt}. Here we performed AT by generating adversarial examples via PGD method with $\epsilon=8/255$ in $l_{\infty}$-norm bound and $T = 2$ iterative steps ($s = 8/255$).

%is submitted along with the paper and will be publicly available.

%\vspace{-1mm}
\subsection{Comparison Results}

\paragraph{The Standard ZSL Setting.}
% We also report the robustness of LAAT against
We performed experiments in the standard and usual 5-way one/zero-shot setting \cite{basefew, AQ, AdvMAML}. Following previous works \cite{AQ, AdvMAML}, we report the robustness of our method against common adversarial attacks, including FGSM \cite{adv_attack2}, PGD \cite{pgdat} with 20 iterative steps and step size $s = 2/255$ (denoted as PGD-20 in the following text), CW \cite{cw} (optimized by PGD for 30 steps with step size $s = 0.8/255$), and AutoAttack (AA) \cite{autoattack}, an effective ensemble attack (several variants of PGD and one query-based black-box Square attack \cite{square}) used to assess adversarial robustness. The maximal perturbation was set to commonly used $\epsilon=8/255$ in $l_{\infty}$ setting. The experimental results are reported from 2000 randomly sampled 5-way tasks. The results presented in \cref{tab:cifarfs} demonstrate that LAAT in the zero-shot setting achieved competitive performance with several methods in the few-shot setting. Notably, the adversarial robustness of LAAT models even surpassed that of the previous SOTA adversarially robust few-shot method GR \cite{basefew}. Additionally, our method exhibited significantly better clean accuracy than all three few-shot methods. We note that these results should be interpreted cautiously because LAAT and these few-shot works used different auxiliary information: the former used text information of the test category names and the latter relied on a small number of example images from the test categories.

\vspace{-2mm}
\paragraph{GZSL Setting across Datasets.}
In the GZSL setting, the models were evaluated in $N$-way classification setting, where $N$ is the number of categories in each dataset. All images on downstream datasets are resized to 224 $\times$ 224 in inference. Here we compared LAAT with TeCoA. TeCoA originally performed adversarial fine-tuning on ImageNet-1K with maximum perturbation $\epsilon = 1/255$. For a fair comparison, here adversarial fine-tuning was also performed with the released code of TeCoA \cite{clip_finetune}. \cref{tab:pretrain} shows the zero-shot results on ten downstream datasets under PGD-20. Here the accuracy is reported by sampling 100 examples per category. We can see that TeCoA almost failed to train adversarially robust models in the common $\epsilon$ setting, while the models trained with LAAT can obtain substantial adversarial robustness on several downstream datasets. 

Note that the models trained with LAAT did not perform quite well on Cifar100, DTD \cite{dtddata}, and SUN \cite{sundata}. This could be caused by the original small resolution of Cifar100 (32 $\times$ 32) \cite{recipt} and the large inter-class differences between DTD, SUN, and ImageNet-1K.

\begin{figure}[t]
  \centering
  \begin{subfigure}[b]{0.49\linewidth}
         \centering
         \includegraphics[width=\linewidth]{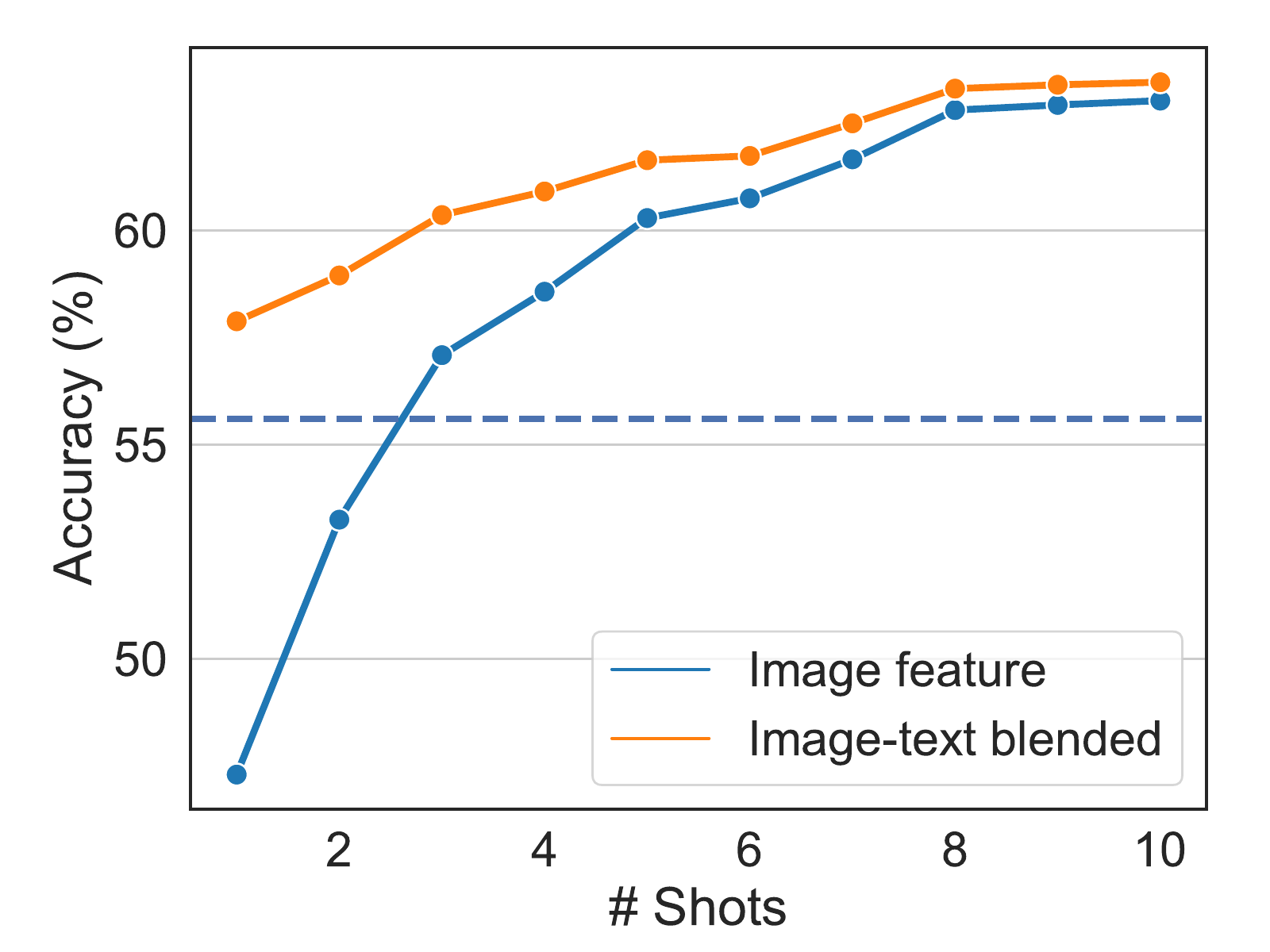}
         \caption{Benign accuracy.}
     \end{subfigure}
     \hfill
     \begin{subfigure}[b]{0.49\linewidth}
         \centering
         \includegraphics[width=\linewidth]{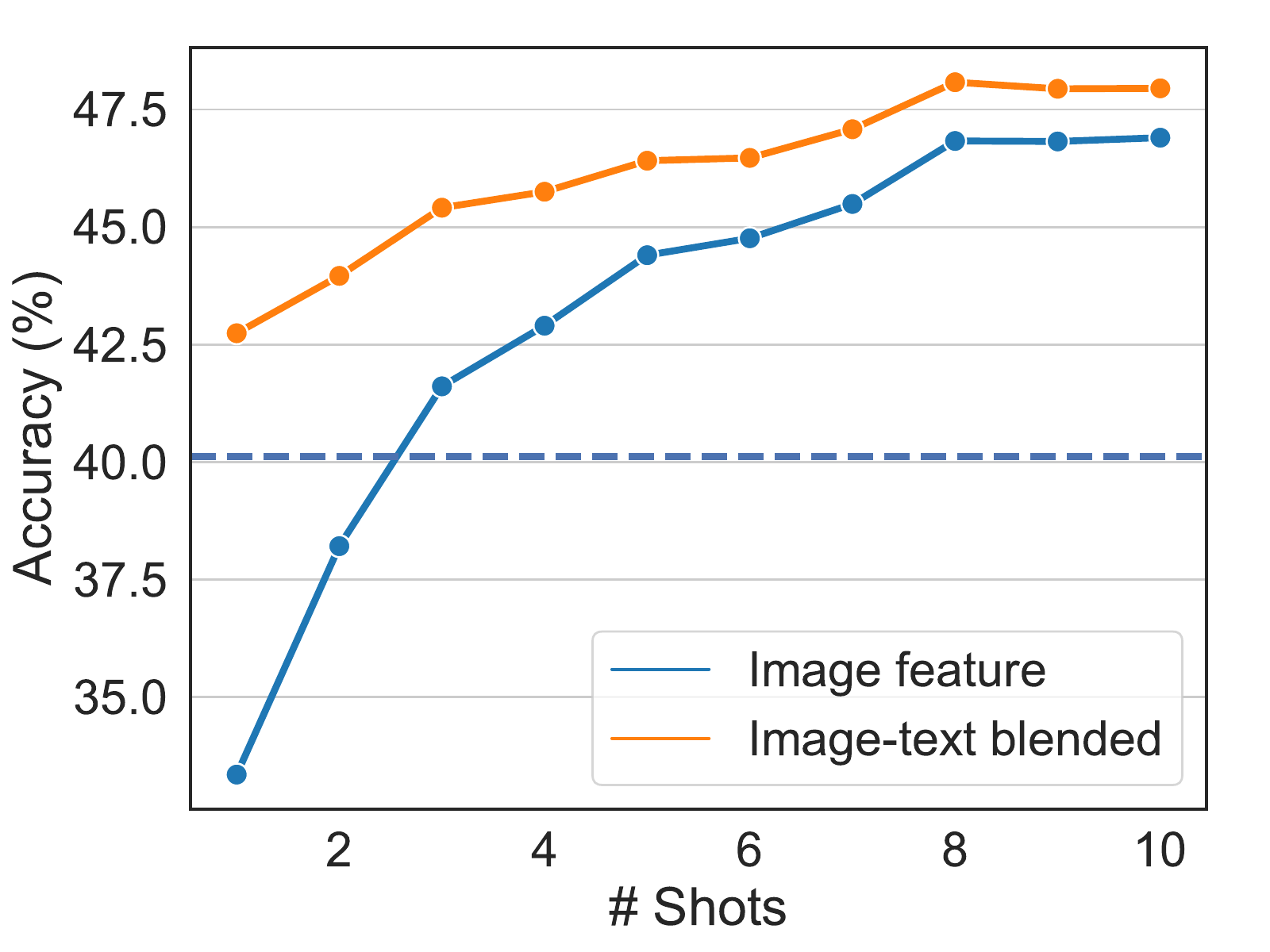}
         \caption{Robust accuracy.}
     \end{subfigure}
%    \vspace{-2mm}
   \caption{Classification accuracy on both benign and adversarial examples in 5-way few-shot setting on CIFAR-FS, with image feature anchors or with image-text blended anchors. The dashed line denotes 5-way zero-shot accuracy.}
   \label{fig:zerotofew}
\end{figure}

%\vspace{-1mm}
\subsection{Extending LAAT to Few-Shot Setting}
\label{sec:ztof}
Our zero-shot models supervised by LAAT can be naturally extended to the few-shot setting and the robustness can be further improved. Given the images of novel categories as the support set, we can follow the previous few-shot SOTA method \cite{basefew} and use the prototype-based metric learning \cite{metric} to build image feature-based anchors. That is, in the $K$-shot case, the image anchor $\mathbf{v}_y$ for a novel category $y$ can be computed as $\mathbf{v}_y = {\rm Norm_2}\{\sum_{i=1}^K f_\Theta(\mathbf{x}_i)\}$, where $\mathbf{x}_i$ denotes the images of $y$ in the support set and $\rm{Norm_2}\{\cdot\}$ denotes the $l_2$ normalization. We can also weight the original text anchor $\mathbf{a}_y$ from the text encoder and the image feature from the support set to build an image-text blended anchor for $y$: $ \mathbf{\hat{a}}^\star_y = {\rm Norm_2}\{{\beta \cdot \mathbf{\hat{a}}_y + \sum_{i=1}^K f_\Theta(\mathbf{x}_i)}\}$. Here the text anchor can be regarded as the \textit{prior knowledge} of $y$. We set $\beta = 2$ in all experiments. 

\cref{tab:cifarfs} shows the results of different models with image-text blended anchors in the 1-shot setting. The 1-shot performance of LAAT was better than the zero-shot performance. \cref{fig:zerotofew} further shows the results with two different anchors $\mathbf{\hat{a}}^\star$ and $\mathbf{v}$ in $K$-shot settings. We can see that the text anchor used to perform zero-shot recognition boosted the performance in various $K$-shot settings, especially when the support set is small.

\begin{table}[!t]
  \centering
  \small
%  \resizebox{\linewidth}{36mm}{
%	\renewcommand{\arraystretch}{1.15}{
%  \setlength{\tabcolsep}{4pt}
  {
    \begin{tabular}{c|cc}
%    \toprule
     Objective & Clean & Robust  \\
%     \midrule
		\hline
     \textit{Eucl} & $39.17$ & $28.04$  \\
     $\theta$ & $37.92$ & $27.84$  \\
     $\cos \theta$ & $41.91$ & $27.91$  \\
%     \bottomrule

    \end{tabular}
    
    }
%   }
%	\vspace{-2mm}
   \caption{Classification accuracy (\%) on CIFAR-FS supervised by the original CLIP anchors with different optimization objectives.}
  \label{tab:difficulty}
\end{table}

\begin{table}[!t]
  \centering
%  \vspace{-2mm}
  \small
%  \resizebox{\linewidth}{36mm}{
%	\renewcommand{\arraystretch}{1.15}{
% \setlength{\tabcolsep}{4pt}
  {
    \begin{tabular}{ccc|cc}
%    \toprule
%    \multicolumn{3}{c||}{Method} & \multicolumn{2}{c}{CIFAR-FS} \\
%    \cline{1-4}
       EXP & A-CE & SM & Clean & Robust  \\
%     \midrule
	\hline
       $$ & $$ & $ $  & $41.91$ & $27.91$ \\
       $\checkmark$ & $$ & $$ & $46.98$ & $32.71$ \\
       $\checkmark$ & $\checkmark$ & $ $ & $54.02$ & $37.69$ \\
       $\checkmark$ & $\checkmark$ & $\checkmark$ & $\textbf{55.60}$ & $\textbf{40.12}$ \\
%     \bottomrule

    \end{tabular}
    
    }
%   }
%	\vspace{-2mm}
   \caption{Classification accuracy (\%) of Conv4-512 on CIFAR-FS in different ablation experiments. The first two lines were the results obtained with $\cos \theta$ as the objective.}
  \label{tab:ablation}
\end{table}

%\vspace{-1mm}
\subsection{Ablation Study}
\label{sec:ablation}
Limited by computing resources, the ablation study is not studied on ImageNet-1K. Unless specified, it was performed with Conv4-512 in the 5-way zero-shot setting. The robustness was evaluated by the PGD-20 attack method.

%the robust accuracy is evaluated by PGD-20.

%\vspace{-2mm}

\vspace{-2mm}
\paragraph{Effectiveness of each design.}
%\textit{Supplementary Materials}. 

In \cref{fig:curve}, we have shown that if the model is supervised by CoS between the original CLIP anchors and the output feature, say, $\cos \theta$, it is difficult to converge under AT. In \cref{tab:difficulty}, we show the results of other anchor-based AT optimization objectives, such as the angle \cite{hyper} and the Euclidean distance \cite{mmc}, denoted as $\theta$ and \textit{Eucl} respectively, in the ZSL setting. The results indicate that none of these three optimization objectives led to good classification accuracy. Note that random guessing in this setting can obtain 20\% accuracy. 

We then conducted an ablation study to show the effectiveness of each design in LAAT: the expansion algorithm, the A-CE loss, and the smoothness loss, denoted as EXP, A-CE, and SM, respectively. The results in \cref{tab:ablation} show that all of these designs are helpful in improving the zero-shot adversarial robustness. 

As stated in \cref{sec:objective}, A-CE could make a relaxation on the original anchor-based AT objectives. To validate this, we computed the average CoS between GT anchors and the output visual features of the two models trained: EXP and EXP + A-CE (EXP was supervised by $\cos \theta$; see \cref{tab:ablation}). With A-CE loss, the visual feature's CoS with GT anchors dropped from 0.455 to 0.205 on average, while the accuracies on both benign and adversarial examples were improved (see \cref{tab:ablation}). These results support our conjecture that some examples not quite close to the GT anchor can still be classified correctly with A-CE supervision.

\vspace{-2mm}
\paragraph{Smoothness v.s. TRADES.}
 It is seen that SM has similar objectives with TRADES \cite{trades}, but they are different in many aspects (see \cref{sec:smoothness_trades}). Here we also compare SM with TRADES empirically. The experiments were performed based on expanded anchors with A-CE supervision. The results are shown in \cref{tab:robustness}, together with the most important hyper-parameter of each method. We can clearly see that TRADES performed much worse than our smoothness loss. In addition, the results show that the effectiveness of smoothing loss was not quite sensitive to $\alpha$.

\begin{table}[!t]
  \centering
  \small
%  \resizebox{\linewidth}{36mm}{
%	\renewcommand{\arraystretch}{1.15}{
 \setlength{\tabcolsep}{3pt}
  {
    \begin{tabular}{c|c|cc}
%    \toprule
     AT Method & Hyper-Parameter & Clean & Robust  \\
     \hline
     \multirow{2}*{EXP + A-CE + TRADES} & $1/\lambda = 1$ & $55.26$ & $19.22$ \\
    %  & $1/\lambda = 3$ &$53.8 \pm 0.5$ &$20.8 \pm 0.4$ \\
      & $1/\lambda = 6$ & $55.82$ & $25.02$ \\
%     \midrule
%     \multirow{2}*{AWP \cite{awp}} & $\gamma = 1e^{-3}$ & $53.49$ & $37.69$  \\
    %  & $\gamma = 5e^{-3}$ &$54.2 \pm 0.5$ &$40.1 \pm 0.4$ \\
%     & $\gamma = 1e^{-2}$ & $\textbf{57.18}$ & $39.41$ \\
     \hline
     \multirow{3}*{EXP + A-CE + SM} & $\alpha = 1.0$ & $55.62$ & $39.95$ \\
     & $\alpha = 3.0$ & $55.60$ & $40.12$ \\
     & $\alpha = 6.0$ & $53.01$ & $40.29$ \\
    %  & $\lambda = 6$ &$54.7 \pm 0.5$ &$42.6 \pm 0.4$ \\
%     \midrule
%     \midrule
%     \multirow{2}*{SM + AWP}& $\gamma = 1e^{-2}, \alpha = 1.0$ & $56.70$ & $40.50$ \\
%                            & $\gamma = 1e^{-2}, \alpha = 3.0$ & $53.22$ & $\textbf{40.73}$ \\
    %  & $\lambda = 6$ &$52.3 \pm 0.5$ &$42.2 \pm 0.4$ \\
     
%     \bottomrule
	
    \end{tabular}
    
    }
%   }
%	\vspace{-2mm}
   \caption{Classification accuracy (\%) on CIFAR-FS with different AT methods.}
  \label{tab:robustness}
\end{table}

\begin{table}[!t]
  \centering
  \small
%  \resizebox{\linewidth}{36mm}{
%	\renewcommand{\arraystretch}{1.15}{
  \setlength{\tabcolsep}{4pt}
  {
    \begin{tabular}{c|c|cc|cc}
%    \toprule
     \multirow{2}*{Method} & \multirow{2}*{Setting} & \multicolumn{2}{c|}{Conv4-64} & \multicolumn{2}{c}{ResNet12} \\
      & & Clean & Robust & Clean & Robust \\
%     \midrule
	\hline
     AQ & 1-shot & $43.74$ & $27.54$ & $43.58$ & $31.45$ \\
     R-MAML & 1-shot & $32.46$ & $24.61$ & $42.54$ & $32.54$ \\
     GR & 1-shot & $44.51$ & $37.45$ &  $48.13$ & $39.29$ \\
%     \midrule
	\hline
     LAAT & zero-shot & $\textbf{49.84}$ & $\textbf{38.74}$ &  $\textbf{58.37}$ & $\textbf{42.88}$  \\
%     LAAT & 1-shot & $\textbf{51.75}$ & $\textbf{40.78}$ &  $\textbf{59.87}$ & $\textbf{44.22}$  \\
     
%     \bottomrule
	
    \end{tabular}
    
    }
%   }
   \caption{Classification accuracy $(\%)$ on CIFAR-FS with different image models.}
  \label{tab:imagemodels}
\end{table}

\begin{table}[!t]
  \centering
  \small
%  \resizebox{\linewidth}{36mm}{
%	\renewcommand{\arraystretch}{1.15}{
%  \setlength{\tabcolsep}{4pt}
  {
    \begin{tabular}{cc|cc}
%    \toprule
     Text Encoder & Embed-dim & Clean & Robust \\
%     \midrule
	\hline
		None (MMC, one-shot) & $512$ & $47.55$ & $32.79$ \\
     RN50x4 & $640$ & $52.28$ & $38.65$  \\
     RN50x16 & $768$ & $53.86$ & $38.02$  \\
     ViT-B/32 & $512$ & $52.68$ & $38.39$  \\
     ViT-B/16 & $512$ & $\textbf{55.60}$ & $\textbf{40.12}$  \\
     ViT-L/14 & $768$ & $52.78$ & $39.48$  \\
     
%     \bottomrule
	
    \end{tabular}
    
    }
%   }
%   \vspace{-2mm}
   \caption{Classification accuracy $(\%)$ supervised by MMC anchors in one-shot setting and the full LAAT with different CLIP text encoders in zero-shot setting on CIFAR-FS. Embed-dim denotes the dimension of the text anchors.}
  \label{tab:textencoder}
\end{table}

\vspace{-2mm}
\paragraph{Different image models.} 
We also performed experiments with other popular image models in the few-shot setting other than Conv4-512, \eg, Conv4-64 \cite{conv64} and ResNet12 \cite{AQ}. \cref{tab:imagemodels} shows the results of the two image models on CIFAR-FS. Conv4-64 is smaller than Conv4-512 while ResNet12 is slightly larger. The models trained with LAAT showed stronger zero-shot adversarial robustness than those few-shot methods regardless of the model size.

\vspace{-2mm}
\paragraph{Different text encoders.} 
Without text encoders, it is impossible for LAAT to achieve zero-shot robustness. In principle, arbitrary text encoders are feasible for LAAT if they have semantic consistency. We show the influence of using different CLIP text encoders in \cref{tab:textencoder}. We also show the results without using any text encoders. Instead, the image model was supervised by the MMC anchors (here zero-shot recognition cannot be performed while one-shot results are reported). We can see that using any text encoders can obtain strong zero-shot robustness, significantly surpassing the few-shot robustness without using text encoders. We also studied the relationship between the distances of different CLIP models' anchors and Cifar100 super-categories. The results in \cref{sec:semantic_consistency_performance} further show that semantic consistency of the text encoder is one of the keys to zero-shot adversarial robustness.

\section{Conclusion and Discussion}
\label{sec:discussion}
In this work, we propose a novel LAAT strategy for adversarially robust image classification in a zero-shot setting, inspired by recent vision-language models. Extensive experiments demonstrate that our method has strong adversarial robustness across different models and several zero-shot settings. We hope our work could encourage more researchers to investigate the robustness in the zero-shot setting, which could be a promising way to improve the scalability and flexibility of adversarially robust models. Limited by the data and computing resources, we cannot perform AT on large-scale image-text pairs directly like CLIP \cite{clip}. We believe it could be another promising way to achieve zero-shot adversarial robustness.

%\vspace{-2mm}
%\paragraph{Limitation.} 
\section*{Acknowledgement}

This work was supported in part by the National Natural Science Foundation of China (Nos. U2341228 and U19B2034). 

%and robustly pre-trained large models based on this setting. 

{
    \small
    \bibliographystyle{ieeenat_fullname}
    \bibliography{main}
}

\clearpage
\setcounter{page}{1}
\maketitlesupplementary
\appendix
\renewcommand{\thefigure}{S\arabic{figure}}
\renewcommand{\thetable}{S\arabic{table}}
\setcounter{table}{0} 
\setcounter{figure}{0}

\setlength{\algomargin}{0.7em}
\SetNlSkip{0.4em}
\SetNlSty{text}{}{}
\begin{algorithm}[t!]
    \SetKwInOut{Return}{Return}
    \SetKwInOut{Input}{Input}
    \SetKwInOut{Output}{Output}
         \caption{Expansion Algorithm}
         \label{algo1}
     \Input{
         The original anchors $\{\mathbf{a}_i\}_{i=1}^{N}$, where $\mathbf{a}_i \in \mathbb{R}^{n}$ and $N$ denotes the number of anchor points. \\
         %mapping $\mathcal{R}$.\\
      }
      \Output{
         The expanded text anchors $\{\hat{\mathbf{a}}_i\}_{i=1}^{N}$. \\
        
      }
      Find a \textit{center}: $\mathbf{v} \leftarrow \frac{\sum_{i=1}^{N} \mathbf{a}_i}{||\sum_{i=1}^{N}{\mathbf{a}_i}||_2}$ \\
      Calculate a rotation matrix $\mathbf{R}$ so that $\mathbf{R}\mathbf{v} = \mathbf{p}$, where $\mathbf{p} = [1, 0, \cdots, 0]$ \\
 %    $\mathbf{x} \leftarrow \frac{\mathbf{x}}{255}$\\
 %    $\mathbf{x_0} \leftarrow \mathbf{x}, \mathbf{r} \leftarrow 0, m \leftarrow 0$\\
     \For{$i = 1 \ to \ N$}{
         $\tilde{\mathbf{a}}_i\leftarrow \mathbf{R}\mathbf{a}_i $ \\
         }
     $\phi _0 \leftarrow \max_{1 \leq i \leq N}{\{\arccos{\tilde{a}_{i}^{(1)}}\}}$, where $\tilde{a}_{i}^{(j)}$ denotes the $j$-th element of $\tilde{\mathbf{a}}_i$\\
     \For{$i = 1 \ \mathrm{to} \ N$}{
         $\tilde{\phi}_{i}^{(1)} \leftarrow \arccos \tilde{a}_{i}^{(1)}$\\
         $\bar{\phi}_{i}^{(1)} \leftarrow  \frac{\pi}{2} \cdot \frac{\tilde{\phi}_{i}^{(1)}}{\phi_0}$\\
         $\bar{a}_{i}^{(1)} \leftarrow \cos{\bar{\phi} _{i}^{(1)}}$\\
         \For{$j = 2 \ \mathrm{to} \ n$}{
             $\bar{a}_{i}^{(j)} \leftarrow \tilde{a}_{i}^{(j)} \cdot \sin{\bar{\phi}_{i}^{(1)}} / \sin{\tilde{\phi}_{i}^{(1)}}$\\
             }
         }
     \For{$i = 1 \ \mathrm{to} \ N$}{
         $\hat{\mathbf{a}}_i = \mathbf{R^T}\bar{\mathbf{a}}_i$
     }
     \Return{$\{\hat{\mathbf{a}}_i\}_{i=1}^{N}$.}
\end{algorithm}

\begin{table}[!t]
   \centering
   \small
 %  \resizebox{\linewidth}{36mm}{
 %	\renewcommand{\arraystretch}{1.15}{
 %  \setlength{\tabcolsep}{4pt}
   {
     \begin{tabular}{c|cc}
%     \toprule
      Hyper-Parameter & Clean & Robust  \\
%      \midrule
	 \hline
      $\tau = 0.03$ &$44.37$ &$32.51$ \\
      $\tau = 0.07$ &$51.43$ &$35.23$ \\
      $\tau = 0.2$ & $58.80$ & $37.52$\\
      $\tau = 0.5$ & $56.07$ &$40.17$ \\
      $\tau = 1.0$ & $55.60$ &$40.12$ \\
      
%      \bottomrule
	
     \end{tabular}
    
     }
 %   }
   \caption{Classification accuracy (\%) of Conv4-512 on CIFAR-FS with different $\tau$.  Here all experiments are performed based on expanded anchors with A-CE loss and smoothness loss supervision. The robust accuracy is evaluated by PGD-20.}
   \label{tab:tau}
 \end{table}

\section{Experimental Details on CIFAR100}
\label{sec:experimental_cifar100}

We used a PreActResNet18 model on Cifar100 to investigate the influence of CLIP anchors on anchor-based AT performance. We used the CLIP ViT-B/16 text encoder. The prompt text was set as ``This is a photo of a \{ \}''. Three types of anchors were evaluated: the original CLIP anchors, the expanded CLIP anchors (see Sec. 3.4), and the MMC anchors \cite{mmc} (the average CoS $<$ 0). The optimization objectives were set to maximize the CoS between output feature $\mathbf{z}$ and different types of GT anchors: $\frac{\mathbf{z}^T\mathbf{a}_y}{||\mathbf{z}||_2}$. 

We adversarially trained three models with these anchors by generating adversarial examples via PGD with maximum adversarial perturbation $\epsilon=8/255$ in $l_{\infty}$-norm bound, with iterative steps $T = 10$ and the step size $2/255$. We evaluated the robustness of the models against PGD \cite{pgdat} with 20 iterative steps and step size $2/255$. The model was optimized for 200 epochs by SGD with an initial learning rate of 0.1, a momentum of 0.9, and a weight decay of $5e^{-4}$. The learning rate was reduced by 10 two times after the 100th and 150th epochs. Fig. 3 shows the learning curves of the three models on Cifar100 test set.

% centers of the Max-Mahalanobis distribution

\section{The Derivation from Rotated Anchor to Expanded Anchor}
\label{sec:derivation_expand_anchor}
For the unit hyper-sphere, $r=1$. According to Eq. (1), the coordinates of the rotated anchor $\tilde{\mathbf{a}}_i$ can be written as
\begin{equation}
\begin{aligned}
&    \tilde{a}_{i}^{(1)} = \cos{\tilde{\phi}_{i}^{(1)}}; \\
& \tilde{a}_{i}^{(2)} = \sin{\tilde{\phi}_{i}^{(1)}}\cos{\tilde{\phi}_{i}^{(2)}}; \\
& ...\\
& \tilde{a}_{i}^{(n)} = \sin{\tilde{\phi}_{i}^{(1)}}\sin{\tilde{\phi}_{i}^{(2)}}...\sin{\tilde{\phi}_{i}^{(n-2)}}\sin{\tilde{\phi}_{i}^{(n-1)}},
\end{aligned}
\end{equation}
where $\tilde{a}_{i}^{(j)}$ denotes the $j$-th element of $\tilde{\mathbf{a}}_i$, and $\tilde{\phi}_{i}^{(j)}$ denotes the $j$-th angular coordinates anchor $\tilde{\mathbf{a}}_i$.
%the angle with the $j$-th dimension coordinate of 
Once fixed $\phi_0$, we expand the polar angle as  $\bar{\phi}_{i}^{(1)} = \frac{\pi}{2} \cdot \frac{\tilde{\phi}_{i}^{(1)}}{\phi_0}$, while angles with other coordinates remain unchanged, \ie, $\bar{\phi}_{i}^{(j)} = \tilde{\phi}_{i}^{(j)}, j=2,...,n$.
Then we can deduce the expressions for the coordinates of the expanded anchor as:

\begin{equation}
\begin{aligned}
\bar{a}_{i}^{(1)} &= \cos{\bar{\phi}_{i}^{(1)}}; \\
\bar{a}_{i}^{(2)} &= \sin{\bar{\phi}_{i}^{(1)}}\cos{\bar{\phi}_{i}^{(2)}}\\
&=(\sin{\tilde{\phi}_{i}^{(1)}}\cos{\tilde{\phi}_{i}^{(2)}})\cdot \sin{\bar{\phi}_{i}^{(1)}} / \sin{\tilde{\phi}_{i}^{(1)}}\\
&=\tilde{a}_{i}^{(2)}\cdot \sin{\bar{\phi}_{i}^{(1)}} / \sin{\tilde{\phi}_{i}^{(1)}} \\
...\\
\bar{a}_{i}^{(n)} &= \sin{\bar{\phi}_{i}^{(1)}}\sin{\bar{\phi}_{i}^{(2)}}...\sin{\bar{\phi}_{i}^{(n-1)}}\\
&=(\sin{\tilde{\phi}_{i}^{(1)}}\sin{\tilde{\phi}_{i}^{(2)}}...\sin{\tilde{\phi}_{i}^{(n-1)}})\cdot \sin{\bar{\phi}_{i}^{(1)}} / \sin{\tilde{\phi}_{i}^{(1)}}\\
&=\tilde{a}_{i}^{(n)}\cdot \sin{\bar{\phi}_{i}^{(1)}} / \sin{\tilde{\phi}_{i}^{(1)}}.
\end{aligned}
\end{equation}
Thus, we the Cartesian coordinates of expanded anchors $\bar{\mathbf{a}}_i$ are given by:
\begin{equation}
\begin{split}
& \bar{a}_{i}^{(1)} = \cos{\bar{\phi}_{i}^{(1)}}; \\
& \bar{a}_{i}^{(j)} = \tilde{a}_{i}^{(j)} \cdot \sin{\bar{\phi}_{i}^{(1)}} / \sin{\tilde{\phi}_{i}^{(1)}}, \quad j = 2,\cdots,n.
\end{split}
\end{equation}

\section{Pseudo Code of Expansion Algorithm}
\label{sec:pseudo_code}

The pseudo-code of the expansion algorithm can be found in Algorithm 1. Given the original anchors $\{\mathbf{a}_i\}_{i=1}^{N}$, we can obtain the expanded final anchors $\{\hat{\mathbf{a}}_i\}_{i=1}^{N}$ by this algorithm.

\section{The Influence of $\tau$}
\label{sec:ablation_tau}

As stated in Sec. 3.5, several previous works \cite{lseg, clip} on standard training used a temperature parameter $\tau = 0.07$ to scale the CoS, \ie, $L_1$ becomes: 
\begin{equation}
\begin{aligned}
	L^{\star}_1 = \mathbb{E}_{(\mathbf{x}, y)}\left[-\log {\frac{\exp(f_\Theta(\mathbf{x} + \mathbf{\delta})^T\hat{\mathbf{a}}_y/\tau)}{\sum_{i=1}^N \exp(f_\Theta(\mathbf{x} + \mathbf{\delta})^T \hat{\mathbf{a}}_i/\tau)}}\right].
\end{aligned}
\end{equation}
We empirically observed that this might be harmful to the zero-shot performance under AT. \cref{tab:tau} shows the classification accuracy of Conv4-512 on CIFAR-FS with different $\tau$ in the 5-way zero-shot setting (similar to those in Tab. 3). We can see that the model with smaller $\tau$ has worse adversarial robustness in the zero-shot setting. Thus, for simplicity, we set $\tau = 1$ in all other experiments.

\section{Introduction to Downstream Datasets}
\label{sec:downsteam_datasets}

We evaluate the zero-shot adversarial robustness trained on ImageNet-1K on ten downstream datasets, covering a diverse range of recognition tasks. AwA2 \cite{awa2} and aPY \cite{aPY} are two popular datasets in the ZSL setting. COCO Objects are images extracted from the bounding box annotations of MS COCO \cite{coco}. We also include Cifar100 \cite{cifar10}, STL10 \cite{stl10}, Caltech101 \cite{caltech101}, and Caltech256 \cite{caltech256} for generic classification; OxfordPet \cite{oxfordpet} for fine-grained classification; DTD \cite{dtddata} for texture recognition; and SUN \cite{sundata} for  scene recognition.

\section{Smoothness v.s. TRADES}
\label{sec:smoothness_trades}

Smoothness in LAAT is different from TRADES \cite{trades} in several aspects. First, TRADES tries to minimize the classification loss on benign examples and the KL divergence between outputs of benign and adversarial examples, while we try to minimize the classification loss on adversarial examples and maximize the Cosine Similarities (CoS) between benign examples and adversarial examples. Second, the adversarial generation of TRADES is derived from the KL divergence, while the adversarial generation of LAAT only uses the classification loss ($L_1$).

\section{Semantic Consistency v.s. Zero-Shot Performance}
\label{sec:semantic_consistency_performance}

We used different CLIP text encoders to investigate the relationship between the semantic consistency of text encoders and the zero-shot adversarial performance. We performed experiments on Cifar100, which has 100 categories belonging to 20 super-categories. Each super-category includes 5 categories, \eg, \textit{apple}, \textit{mushroom}, \textit{sweet pepper}, \textit{orange}, and \textit{pear} belong to the same super-category \textit{fruit and vegetables}. The categories within the same super-category are semantically similar categories. Therefore, a text encoder with high semantic consistency should map the categories within the same super-category to neighboring anchors and these anchors should have high CoS. We designed two metrics to measure the semantic consistency of a text encoder by using the super-categories.
% of one anchor (with all the 100 anchors)

We first calculated CoS between the category anchors obtained from a text encoder and then numbered the categories in descending order of the CoS. The numbers ranged between 0 to 99. We name them as \textit{ranks} next. \cref{ViT-B/16} shows the ranks of five categories in the super-category \textit{fruit and vegetables} with text encoder ViT-B/16. For example, the \textit{apple} anchor's CoS with the \textit{pear} anchor is the 2nd highest among its 99 CoS with other anchors (except the \textit{apple} anchor itself), so the rank in the first row and the fifth column of \cref{ViT-B/16} is 2.

\begin{table}[!t]
  \centering
  \small
 \setlength{\tabcolsep}{2.5pt}
  {
    \begin{tabular}{c|ccccc}
    \toprule
     Category & Apple & Mushroom & Sweet pepper &  Orange &  Pear \\
     \midrule
     Apple & $0$ & $7$ & $1$ & $27$ & $2$ \\ 
     Mushroom & $7$ & $0$ & $4$ & $23$ & $34$ \\
     Sweet pepper & $1$ & $11$ & $0$ & $32$ & $10$ \\
     Orange & $4$ & $10$ & $3$ & $0$ & $2$ \\
     Pear & $2$ & $30$ & $3$ & $6$ & $0$ \\
     \bottomrule
    \end{tabular}
    
    }
   \caption{The ranks of categories belonging to the super-category \textit{fruit and vegetables} with text encoder ViT-B/16. Each rank is calculated with the row category and all column categories. Each category has the highest CoS with itself, so the ranks on the diagonal are always 0.}
  \label{ViT-B/16}
\end{table}

% Note that the same rank may occur.

\begin{table}[!t]
  \centering
  \small
  {
    \begin{tabular}{c|cc}
    \toprule
    Text Encoder &  Sum of rank & Top-5 ratio\\
     \midrule
    RN50x4 & 332 & 54.6\% \\
    RN50x16 & 319 & 54.4\% \\
    ViT-B/32 & 286 & 60.6\% \\
    ViT-B/16 & \textbf{225} & \textbf{61.4\%} \\
    ViT-L/14 & 284 & 58.2\% \\
     \bottomrule
    \end{tabular}
    
    }
   \caption{The average sum of ranks and the average top-5 ratio of ranks on 20 super-categories for each text encoder. The smaller sum of ranks and the larger top-5 ratio of ranks indicate greater semantic consistency of a text encoder.}
   \label{average}
\end{table}

The two metrics were designed based on the ranks of the CoS described above. Intuitively, the higher the CoS within a super-category, the smaller the ranks within this super-category. Note that using ranks has an advantage over using the CoS directly, as ranks take the relationships between different super-categories into consideration. We calculated the sum of ranks and top-5 ratio of ranks within each super-category, and averaged them over all 20 super-categories as the two metrics. Take \textit{fruit and vegetables} in \cref{ViT-B/16} as an example, the former is the sum of $5 \times 5 = 25$ ranks, 219 in this case, and the latter is the ratio of ranks from 0 to 4 out of 25 ranks, $14/25 = 56\%$ in this case. The smaller sum of ranks and the larger top-5 ratio of ranks indicate greater semantic consistency of a text encoder. 

\cref{average} shows each text encoder's average sum of ranks and the average top-5 ratio of ranks on the 20 super-categories. We can see that ViT models are better than ResNet models under these two metrics. These results on semantic consistency generally correspond to the results on zero-shot adversarial robustness in Tab. 7 (note that CIFAR-FS \cite{cifarfs} used in Tab. 7 is a variant of Cifar100), \eg, ViT-B/16 has the best semantic consistency under these two metrics and it also has the best zero-shot adversarial robustness among these models. This correspondence indicates that the semantic consistency of the text encoder is one of the keys to zero-shot adversarial robustness.

\end{document}

%% file: preamble.tex
\usepackage[dvipsnames]{xcolor}

\def\eg{\emph{e.g.}}

\def\ie{\emph{i.e.}}

%% file: main.bbl
\begin{thebibliography}{57}
\providecommand{\natexlab}[1]{#1}
\providecommand{\url}[1]{\texttt{#1}}
\expandafter\ifx\csname urlstyle\endcsname\relax
  \providecommand{\doi}[1]{doi: #1}\else
  \providecommand{\doi}{doi: \begingroup \urlstyle{rm}\Url}\fi

\bibitem[Aguilera and P{\'{e}}rez{-}Aguila(2004)]{rotate}
Antonio Aguilera and Ricardo P{\'{e}}rez{-}Aguila.
\newblock General n-dimensional rotations.
\newblock In \emph{WSCG}, pages 1--8, 2004.

\bibitem[Andriushchenko et~al.(2020)Andriushchenko, Croce, Flammarion, and
  Hein]{square}
Maksym Andriushchenko, Francesco Croce, Nicolas Flammarion, and Matthias Hein.
\newblock Square attack: {A} query-efficient black-box adversarial attack via
  random search.
\newblock In \emph{Eur. Conf. Comput. Vis. (ECCV)}, pages 484--501, 2020.

\bibitem[Athalye et~al.(2018)]{obs}
Anish Athalye et~al.
\newblock Obfuscated gradients give a false sense of security: Circumventing
  defenses to adversarial examples.
\newblock In \emph{Int. Conf. Mach. Learn. (ICML)}, pages 274--283, 2018.

\bibitem[Ban and Dong(2022)]{clip_robust2}
Yuanhao Ban and Yinpeng Dong.
\newblock Pre-trained adversarial perturbations.
\newblock \emph{Adv. Neural Inform. Process. Syst. (NeurIPS)}, 2022.

\bibitem[Bertinetto et~al.(2019)Bertinetto, Henriques, Torr, and
  Vedaldi]{cifarfs}
Luca Bertinetto, Jo{\~{a}}o~F. Henriques, Philip H.~S. Torr, and Andrea
  Vedaldi.
\newblock Meta-learning with differentiable closed-form solvers.
\newblock In \emph{Int. Conf. Learn. Represent. (ICLR)}, 2019.

\bibitem[Blumenson(1960)]{coord}
LE Blumenson.
\newblock A derivation of n-dimensional spherical coordinates.
\newblock \emph{The American Mathematical Monthly}, 67\penalty0 (1):\penalty0
  63--66, 1960.

\bibitem[Carlini and Wagner(2017)]{cw}
Nicholas Carlini and David~A. Wagner.
\newblock Towards evaluating the robustness of neural networks.
\newblock In \emph{{IEEE} Symposium on Security and Privacy, {SP}}, pages
  39--57, 2017.

\bibitem[Chao et~al.(2016)Chao, Changpinyo, Gong, and Sha]{gzsl}
Wei{-}Lun Chao, Soravit Changpinyo, Boqing Gong, and Fei Sha.
\newblock An empirical study and analysis of generalized zero-shot learning for
  object recognition in the wild.
\newblock In \emph{Eur. Conf. Comput. Vis. (ECCV)}, pages 52--68, 2016.

\bibitem[Cimpoi et~al.(2014)Cimpoi, Maji, Kokkinos, Mohamed, and
  Vedaldi]{dtddata}
Mircea Cimpoi, Subhransu Maji, Iasonas Kokkinos, Sammy Mohamed, and Andrea
  Vedaldi.
\newblock Describing textures in the wild.
\newblock In \emph{IEEE Conf. Comput. Vis. Pattern Recog. (CVPR)}, pages
  3606--3613, 2014.

\bibitem[Coates et~al.(2011)Coates, Ng, and Lee]{stl10}
Adam Coates, Andrew Ng, and Honglak Lee.
\newblock An analysis of single-layer networks in unsupervised feature
  learning.
\newblock In \emph{AISTATS}, pages 215--223, 2011.

\bibitem[Croce and Hein(2020)]{autoattack}
Francesco Croce and Matthias Hein.
\newblock Reliable evaluation of adversarial robustness with an ensemble of
  diverse parameter-free attacks.
\newblock In \emph{Int. Conf. Mach. Learn. (ICML)}, pages 2206--2216, 2020.

\bibitem[Debenedetti et~al.(2022)]{recipt}
Edoardo Debenedetti et~al.
\newblock A light recipe to train robust vision transformers.
\newblock \emph{arXiv preprint arXiv:2209.07399}, 2022.

\bibitem[Dong et~al.(2022)Dong, Wang, Lai, and Xie]{basefew}
Junhao Dong, Yuan Wang, Jianhuang Lai, and Xiaohua Xie.
\newblock Improving adversarially robust few-shot image classification with
  generalizable representations.
\newblock In \emph{IEEE Conf. Comput. Vis. Pattern Recog. (CVPR)}, pages
  9015--9024. {IEEE}, 2022.

\bibitem[Dosovitskiy et~al.(2021)Dosovitskiy, Beyer, Kolesnikov,
  et~al.]{transformer}
Alexey Dosovitskiy, Lucas Beyer, Alexander Kolesnikov, et~al.
\newblock An image is worth 16x16 words: Transformers for image recognition at
  scale.
\newblock In \emph{Int. Conf. Learn. Represent. (ICLR)}, 2021.

\bibitem[Duan et~al.(2022)Duan, Kang, Wang, Han, Xue, Wang, and
  Zhang]{rethinkingmaml}
Xiaoyue Duan, Guoliang Kang, Runqi Wang, Shumin Han, Song Xue, Tian Wang, and
  Baochang Zhang.
\newblock Rethinking the number of shots in robust model-agnostic
  meta-learning.
\newblock \emph{arXiv preprint arXiv:2211.15180}, 2022.

\bibitem[Eykholt et~al.(2018)Eykholt, Evtimov, Fernandes, Li, Rahmati, Xiao,
  Prakash, Kohno, and Song]{phy_adv1}
Kevin Eykholt, Ivan Evtimov, Earlence Fernandes, Bo Li, Amir Rahmati, Chaowei
  Xiao, Atul Prakash, Tadayoshi Kohno, and Dawn Song.
\newblock Robust physical-world attacks on deep learning visual classification.
\newblock In \emph{IEEE Conf. Comput. Vis. Pattern Recog. (CVPR)}, pages
  1625--1634, 2018.

\bibitem[Farhadi et~al.(2009)Farhadi, Endres, Hoiem, and Forsyth]{aPY}
Ali Farhadi, Ian Endres, Derek Hoiem, and David~A. Forsyth.
\newblock Describing objects by their attributes.
\newblock In \emph{IEEE Conf. Comput. Vis. Pattern Recog. (CVPR)}, pages
  1778--1785, 2009.

\bibitem[Fei-Fei et~al.(2004)Fei-Fei, Fergus, and Perona]{caltech101}
Li Fei-Fei, Rob Fergus, and Pietro Perona.
\newblock Learning generative visual models from few training examples: An
  incremental bayesian approach tested on 101 object categories.
\newblock In \emph{CVPRW}, pages 178--178, 2004.

\bibitem[Goldblum et~al.(2020)]{AQ}
Micah Goldblum et~al.
\newblock Adversarially robust few-shot learning: {A} meta-learning approach.
\newblock In \emph{Adv. Neural Inform. Process. Syst. (NeurIPS)}, 2020.

\bibitem[Goodfellow et~al.(2015)]{adv_attack2}
Ian~J. Goodfellow et~al.
\newblock Explaining and harnessing adversarial examples.
\newblock In \emph{Int. Conf. Learn. Represent. (ICLR)}, 2015.

\bibitem[Griffin et~al.(2007)Griffin, Holub, and Perona]{caltech256}
Gregory Griffin, Alex Holub, and Pietro Perona.
\newblock Caltech-256 object category dataset.
\newblock 2007.

\bibitem[Gu et~al.(2021)Gu, Lin, Kuo, and Cui]{vild}
Xiuye Gu, Tsung{-}Yi Lin, Weicheng Kuo, and Yin Cui.
\newblock Zero-shot detection via vision and language knowledge distillation.
\newblock In \emph{Int. Conf. Learn. Represent. (ICLR)}, 2021.

\bibitem[He et~al.(2016)He, Zhang, Ren, and Sun]{resnet12}
Kaiming He, Xiangyu Zhang, Shaoqing Ren, and Jian Sun.
\newblock Deep residual learning for image recognition.
\newblock In \emph{IEEE Conf. Comput. Vis. Pattern Recog. (CVPR)}, pages
  770--778, 2016.

\bibitem[Jia et~al.(2021)Jia, Yang, Xia, Chen, Parekh, Pham, Le, Sung, Li, and
  Duerig]{noises}
Chao Jia, Yinfei Yang, Ye Xia, Yi{-}Ting Chen, Zarana Parekh, Hieu Pham,
  Quoc~V. Le, Yun{-}Hsuan Sung, Zhen Li, and Tom Duerig.
\newblock Scaling up visual and vision-language representation learning with
  noisy text supervision.
\newblock In \emph{Int. Conf. Mach. Learn. (ICML)}, pages 4904--4916, 2021.

\bibitem[Krizhevsky et~al.(2009)Krizhevsky, Hinton, et~al.]{cifar10}
Alex Krizhevsky, Geoffrey Hinton, et~al.
\newblock Learning multiple layers of features from tiny images.
\newblock \emph{California Institute of Technology}, 2009.

\bibitem[Kuang et~al.(2023)Kuang, Liu, Wu, and Ji]{semantically}
Huafeng Kuang, Hong Liu, Yongjian Wu, and Rongrong Ji.
\newblock Semantically consistent visual representation for adversarial
  robustness.
\newblock \emph{IEEE Transactions on Information Forensics and Security}, 2023.

\bibitem[Lampert et~al.(2014)Lampert, Nickisch, and Harmeling]{zero_shot_e}
Christoph~H. Lampert, Hannes Nickisch, and Stefan Harmeling.
\newblock Attribute-based classification for zero-shot visual object
  categorization.
\newblock \emph{IEEE Trans. Pattern Anal. Mach. Intell. (TPAMI)}, 36\penalty0
  (3):\penalty0 453--465, 2014.

\bibitem[Li et~al.(2022)Li, Weinberger, Belongie, Koltun, and Ranftl]{lseg}
Boyi Li, Kilian~Q. Weinberger, Serge~J. Belongie, Vladlen Koltun, and
  Ren{\'{e}} Ranftl.
\newblock Language-driven semantic segmentation.
\newblock In \emph{Int. Conf. Learn. Represent. (ICLR)}, 2022.

\bibitem[Li et~al.(2023{\natexlab{a}})Li, Chen, and Hu]{li2023importance}
Xiao Li, Hang Chen, and Xiaolin Hu.
\newblock On the importance of backbone to the adversarial robustness of object
  detectors.
\newblock \emph{arXiv preprint arXiv:2305.17438}, 2023{\natexlab{a}}.

\bibitem[Li et~al.(2023{\natexlab{b}})Li, Wang, Zhang, Sun, and Hu]{rock}
Xiao Li, Ziqi Wang, Bo Zhang, Fuchun Sun, and Xiaolin Hu.
\newblock Recognizing object by components with human prior knowledge enhances
  adversarial robustness of deep neural networks.
\newblock \emph{IEEE Trans. Pattern Anal. Mach. Intell. (TPAMI)},
  2023{\natexlab{b}}.

\bibitem[Lin et~al.(2014)Lin, Maire, Belongie, Hays, Perona, Ramanan,
  Doll{\'{a}}r, and Zitnick]{coco}
Tsung{-}Yi Lin, Michael Maire, Serge~J. Belongie, James Hays, Pietro Perona,
  Deva Ramanan, Piotr Doll{\'{a}}r, and C.~Lawrence Zitnick.
\newblock Microsoft {COCO:} common objects in context.
\newblock In \emph{Eur. Conf. Comput. Vis. (ECCV)}, pages 740--755, 2014.

\bibitem[Madry et~al.(2018)Madry, Makelov, Schmidt, Tsipras, and Vladu]{pgdat}
Aleksander Madry, Aleksandar Makelov, Ludwig Schmidt, Dimitris Tsipras, and
  Adrian Vladu.
\newblock Towards deep learning models resistant to adversarial attacks.
\newblock In \emph{Int. Conf. Learn. Represent. (ICLR)}, 2018.

\bibitem[Mao et~al.(2019)Mao, Zhong, Yang, Vondrick, and Ray]{mao2019metric}
Chengzhi Mao, Ziyuan Zhong, Junfeng Yang, Carl Vondrick, and Baishakhi Ray.
\newblock Metric learning for adversarial robustness.
\newblock \emph{Adv. Neural Inform. Process. Syst. (NeurIPS)}, 32, 2019.

\bibitem[Mao et~al.(2023)Mao, Geng, Yang, Wang, and Vondrick]{clip_finetune}
Chengzhi Mao, Scott Geng, Junfeng Yang, Xin Wang, and Carl Vondrick.
\newblock Understanding zero-shot adversarial robustness for large-scale
  models.
\newblock \emph{Int. Conf. Learn. Represent. (ICLR)}, 2023.

\bibitem[Oord et~al.(2018)Oord, Li, and Vinyals]{infonce}
Aaron van~den Oord, Yazhe Li, and Oriol Vinyals.
\newblock Representation learning with contrastive predictive coding.
\newblock \emph{arXiv preprint arXiv:1807.03748}, 2018.

\bibitem[Pang et~al.(2020{\natexlab{a}})Pang, Xu, Dong, Du, Chen, and Zhu]{mmc}
Tianyu Pang, Kun Xu, Yinpeng Dong, Chao Du, Ning Chen, and Jun Zhu.
\newblock Rethinking softmax cross-entropy loss for adversarial robustness.
\newblock In \emph{Int. Conf. Learn. Represent. (ICLR)}, 2020{\natexlab{a}}.

\bibitem[Pang et~al.(2020{\natexlab{b}})Pang, Yang, Dong, Xu, Zhu, and
  Su]{hyper}
Tianyu Pang, Xiao Yang, Yinpeng Dong, Taufik Xu, Jun Zhu, and Hang Su.
\newblock Boosting adversarial training with hypersphere embedding.
\newblock In \emph{Adv. Neural Inform. Process. Syst. (NeurIPS)},
  2020{\natexlab{b}}.

\bibitem[Parkhi et~al.(2012)Parkhi, Vedaldi, Zisserman, and Jawahar]{oxfordpet}
Omkar~M Parkhi, Andrea Vedaldi, Andrew Zisserman, and CV Jawahar.
\newblock Cats and dogs.
\newblock In \emph{IEEE Conf. Comput. Vis. Pattern Recog. (CVPR)}, pages
  3498--3505, 2012.

\bibitem[Radford et~al.(2021)Radford, Kim, Hallacy, Ramesh, Goh, Agarwal,
  Sastry, et~al.]{clip}
Alec Radford, Jong~Wook Kim, Chris Hallacy, Aditya Ramesh, Gabriel Goh,
  Sandhini Agarwal, Girish Sastry, et~al.
\newblock Learning transferable visual models from natural language
  supervision.
\newblock In \emph{Int. Conf. Mach. Learn. (ICML)}, pages 8748--8763, 2021.

\bibitem[Russakovsky et~al.(2015)Russakovsky, Deng, Su, Krause, Satheesh, Ma,
  Huang, Karpathy, Khosla, Bernstein, Berg, and Fei{-}Fei]{imagenet}
Olga Russakovsky, Jia Deng, Hao Su, Jonathan Krause, Sanjeev Satheesh, Sean Ma,
  Zhiheng Huang, Andrej Karpathy, Aditya Khosla, Michael~S. Bernstein,
  Alexander~C. Berg, and Li Fei{-}Fei.
\newblock Imagenet large scale visual recognition challenge.
\newblock \emph{Int. J. Comput. Vis. (IJCV)}, 115\penalty0 (3):\penalty0
  211--252, 2015.

\bibitem[Snell et~al.(2017)Snell, Swersky, and Zemel]{metric}
Jake Snell, Kevin Swersky, and Richard~S. Zemel.
\newblock Prototypical networks for few-shot learning.
\newblock In \emph{Adv. Neural Inform. Process. Syst. (NeurIPS)}, pages
  4077--4087, 2017.

\bibitem[Szegedy et~al.(2014)Szegedy, Zaremba, Sutskever, Bruna, Erhan,
  Goodfellow, and Fergus]{adv_attack1}
Christian Szegedy, Wojciech Zaremba, Ilya Sutskever, Joan Bruna, Dumitru Erhan,
  Ian~J. Goodfellow, and Rob Fergus.
\newblock Intriguing properties of neural networks.
\newblock In \emph{Int. Conf. Learn. Represent. (ICLR)}, 2014.

\bibitem[Vinyals et~al.(2016{\natexlab{a}})Vinyals, Blundell, Lillicrap,
  Kavukcuoglu, and Wierstra]{MiniImageNet}
Oriol Vinyals, Charles Blundell, Tim Lillicrap, Koray Kavukcuoglu, and Daan
  Wierstra.
\newblock Matching networks for one shot learning.
\newblock In \emph{Adv. Neural Inform. Process. Syst. (NeurIPS)}, pages
  3630--3638, 2016{\natexlab{a}}.

\bibitem[Vinyals et~al.(2016{\natexlab{b}})Vinyals, Blundell, Lillicrap,
  Kavukcuoglu, and Wierstra]{conv64}
Oriol Vinyals, Charles Blundell, Tim Lillicrap, Koray Kavukcuoglu, and Daan
  Wierstra.
\newblock Matching networks for one shot learning.
\newblock In \emph{Adv. Neural Inform. Process. Syst. (NeurIPS)}, pages
  3630--3638, 2016{\natexlab{b}}.

\bibitem[Wang et~al.(2021{\natexlab{a}})]{actionclip}
Mengmeng Wang et~al.
\newblock Actionclip: {A} new paradigm for video action recognition.
\newblock \emph{arXiv preprint arXiv:2109.08472}, 2021{\natexlab{a}}.

\bibitem[Wang et~al.(2021{\natexlab{b}})Wang, Xu, Liu, Chen, Weng, Gan, and
  Wang]{AdvMAML}
Ren Wang, Kaidi Xu, Sijia Liu, Pin{-}Yu Chen, Tsui{-}Wei Weng, Chuang Gan, and
  Meng Wang.
\newblock On fast adversarial robustness adaptation in model-agnostic
  meta-learning.
\newblock In \emph{Int. Conf. Learn. Represent. (ICLR)}, 2021{\natexlab{b}}.

\bibitem[Xian et~al.(2016)Xian, Akata, Sharma, Nguyen, Hein, and
  Schiele]{zla_3}
Yongqin Xian, Zeynep Akata, Gaurav Sharma, Quynh Nguyen, Matthias Hein, and
  Bernt Schiele.
\newblock Latent embeddings for zero-shot classification.
\newblock In \emph{IEEE Conf. Comput. Vis. Pattern Recog. (CVPR)}, pages
  69--77, 2016.

\bibitem[Xian et~al.(2019)Xian, Lampert, Schiele, and Akata]{awa2}
Yongqin Xian, Christoph~H. Lampert, Bernt Schiele, and Zeynep Akata.
\newblock Zero-shot learning - {A} comprehensive evaluation of the good, the
  bad and the ugly.
\newblock \emph{IEEE Trans. Pattern Anal. Mach. Intell. (TPAMI)}, 41\penalty0
  (9):\penalty0 2251--2265, 2019.

\bibitem[Xiao et~al.(2010)Xiao, Hays, Ehinger, Oliva, and Torralba]{sundata}
Jianxiong Xiao, James Hays, Krista~A Ehinger, Aude Oliva, and Antonio Torralba.
\newblock Sun database: Large-scale scene recognition from abbey to zoo.
\newblock In \emph{IEEE Conf. Comput. Vis. Pattern Recog. (CVPR)}, pages
  3485--3492, 2010.

\bibitem[Yu et~al.(2018)Yu, Ji, Fu, Guo, Pang, and Zhang]{zla_2}
Yunlong Yu, Zhong Ji, Yanwei Fu, Jichang Guo, Yanwei Pang, and Zhongfei~(Mark)
  Zhang.
\newblock Stacked semantics-guided attention model for fine-grained zero-shot
  learning.
\newblock In \emph{Adv. Neural Inform. Process. Syst. (NeurIPS)}, pages
  5998--6007, 2018.

\bibitem[Yucel et~al.(2022)]{zerorobust}
Mehmet~Kerim Yucel et~al.
\newblock How robust are discriminatively trained zero-shot learning models?
\newblock \emph{Image Vis. Comput.}, 119:\penalty0 104392, 2022.

\bibitem[Zhang et~al.(2019{\natexlab{a}})Zhang, Yu, Jiao, Xing, Ghaoui, and
  Jordan]{trades}
Hongyang Zhang, Yaodong Yu, Jiantao Jiao, Eric~P. Xing, Laurent~El Ghaoui, and
  Michael~I. Jordan.
\newblock Theoretically principled trade-off between robustness and accuracy.
\newblock In \emph{Int. Conf. Mach. Learn. (ICML)}, pages 7472--7482,
  2019{\natexlab{a}}.

\bibitem[Zhang et~al.(2022)Zhang, Yi, and Sang]{clip_robust3}
Jiaming Zhang, Qi Yi, and Jitao Sang.
\newblock Towards adversarial attack on vision-language pre-training models.
\newblock In \emph{ACM Int. Conf. Multimedia}, pages 5005--5013, 2022.

\bibitem[Zhang et~al.(2021)Zhang, Zhang, Lu, Xiang, Ding, and Huang]{conv4_512}
Manli Zhang, Jianhong Zhang, Zhiwu Lu, Tao Xiang, Mingyu Ding, and Songfang
  Huang.
\newblock {IEPT:} instance-level and episode-level pretext tasks for few-shot
  learning.
\newblock In \emph{Int. Conf. Learn. Represent. (ICLR)}, 2021.

\bibitem[Zhang et~al.(2019{\natexlab{b}})Zhang, Gui, Zhu, Zhao, and
  Liu]{zeroshot_at}
Xingxing Zhang, Shupeng Gui, Zhenfeng Zhu, Yao Zhao, and Ji Liu.
\newblock {ATZSL:} defensive zero-shot recognition in the presence of
  adversaries.
\newblock \emph{arXiv preprint arXiv:1910.10994}, 2019{\natexlab{b}}.

\bibitem[Zhou and Patel(2022)]{zhou2022enhancing}
Mo Zhou and Vishal~M Patel.
\newblock Enhancing adversarial robustness for deep metric learning.
\newblock In \emph{IEEE Conf. Comput. Vis. Pattern Recog. (CVPR)}, pages
  15325--15334, 2022.

\bibitem[Zhu et~al.(2021)Zhu, Li, Li, Wang, and Hu]{phy_adv2}
Xiaopei Zhu, Xiao Li, Jianmin Li, Zheyao Wang, and Xiaolin Hu.
\newblock Fooling thermal infrared pedestrian detectors in real world using
  small bulbs.
\newblock In \emph{AAAI}, pages 3616--3624, 2021.

\end{thebibliography}
